\documentclass[letterpaper]{article} % DO NOT CHANGE THIS
\usepackage{aaai2026}  % DO NOT CHANGE THIS
\usepackage{times}  % DO NOT CHANGE THIS
\usepackage{helvet}  % DO NOT CHANGE THIS
\usepackage{courier}  % DO NOT CHANGE THIS
\usepackage[hyphens]{url}  % DO NOT CHANGE THIS
\usepackage{graphicx} % DO NOT CHANGE THIS
\usepackage{natbib}  % DO NOT CHANGE THIS AND DO NOT ADD ANY OPTIONS TO IT
\usepackage{caption} % DO NOT CHANGE THIS AND DO NOT ADD ANY OPTIONS TO IT
\usepackage{algorithm}
\usepackage{algorithmic}
\usepackage{amsmath}
\usepackage{multirow}
\usepackage{booktabs}
\usepackage{colortbl}

\usepackage{newfloat}
\usepackage{listings}
\DeclareCaptionStyle{ruled}{labelfont=normalfont,labelsep=colon,strut=off} % DO NOT CHANGE THIS
\floatstyle{ruled}
\newfloat{listing}{tb}{lst}{}
\floatname{listing}{Listing}
\nocopyright

\title{FlexQ: Efficient Post-training INT6 Quantization for LLM Serving via Algorithm-System Co-Design}
\author{
    Hao Zhang \textsuperscript{\rm 1}, 
    Aining Jia \textsuperscript{\rm 1}, 
    Weifeng Bu \textsuperscript{\rm 1},
    Yushu Cai \textsuperscript{\rm 1}, 
    Kai Sheng \textsuperscript{\rm 1}, 
    Hao Chen \textsuperscript{\rm 2},
    Xin He \textsuperscript{\rm 1}\thanks{Corresponding author.}
}
\affiliations{
    \textsuperscript{\rm 1} Guangzhou Institute of Technology, Xidian University, 
    \textsuperscript{\rm 2} CSEE, Hunan University\\
    \{zhanghao01, weifengbu, yushucai\}@stu.xidian.edu.cn\\
    jiaaining701@gmail.com, \{kaisheng, hexin\}@xidian.edu.cn, haochen@hnu.edu.cn
}

\begin{document}

\maketitle

\begin{abstract}
Large Language Models (LLMs) demonstrate exceptional performance but entail significant memory and computational costs, restricting their practical deployment. While existing INT4/INT8 quantization reduces these costs, they often degrade accuracy or lack optimal efficiency. INT6 quantization offers a superior trade-off between model accuracy and inference efficiency, but lacks hardware support in modern GPUs, forcing emulation via higher-precision arithmetic units that limit acceleration. 

In this paper, we propose FlexQ, a novel post-training INT6 quantization framework combining algorithmic innovation with system-level optimizations. FlexQ employs uniform 6-bit weight quantization across all layers, with adaptive retention of 8-bit activations in layers identified through layer-wise sensitivity analysis. To maximize hardware efficiency, we develop a specialized high-performance GPU kernel supporting matrix multiplication for W6A6 and W6A8 representations via Binary Tensor Core (BTC) equivalents, effectively bypassing the lack of native INT6 tensor cores.
% Evaluations on LLaMA models show FlexQ maintains near-FP16 accuracy, with perplexity increases of no more than 0.05.
Evaluations on LLaMA family models show FlexQ maintains near-FP16 accuracy, with perplexity increases of no more than 0.1 on WikiText2.
The proposed kernel achieves an average 1.39$\times$ speedup over ABQ-LLM on LLaMA-2-70B linear layers. End-to-end, FlexQ delivers 1.33$\times$ inference acceleration and 1.21$\times$ memory savings over SmoothQuant. Code is released at \url{https://github.com/FlyFoxPlayer/FlexQ}. 
\end{abstract}

\section{Introduction}

Large Language Models (LLMs) have profoundly advanced natural language processing, enabling a wide range of applications from conversational agents to complex reasoning systems \cite{zhao2023survey, kaddour2023challenges, Llama4, guo2025deepseek, comanici2025gemini}. Despite their transformative capabilities, the substantial computational and memory requirements pose significant barriers to their widespread adoption in many commercial settings.

Post-training quantization (PTQ) emerges as a critical technique for accelerating LLM inference. By compressing weights and activations from high-precision formats (e.g., FP32) to low-bit-width representations (e.g., INT8, INT4), PTQ substantially reduces both computational complexity and storage footprint, making model deployment more feasible and efficient for resource-constrained environments. Recently, numerous studies \cite{frantar2022gptq, zhao2024atom, liu2025comet} have pushed toward lower-bit quantization schemes. Despite these advances, the inherent limitations of finite bitwidth representations inevitably introduce accuracy degradation, particularly as the bitwidth decreases. As illustrated in Table \ref{tab:comparison_between_6bit_and_4bit}, current methods maintain relatively stable accuracy at 6-bit precision. However, reducing to 4-bit precision results in a notable performance drop. 

\begin{figure}[h]
    \centering
    \includegraphics[width=\linewidth]{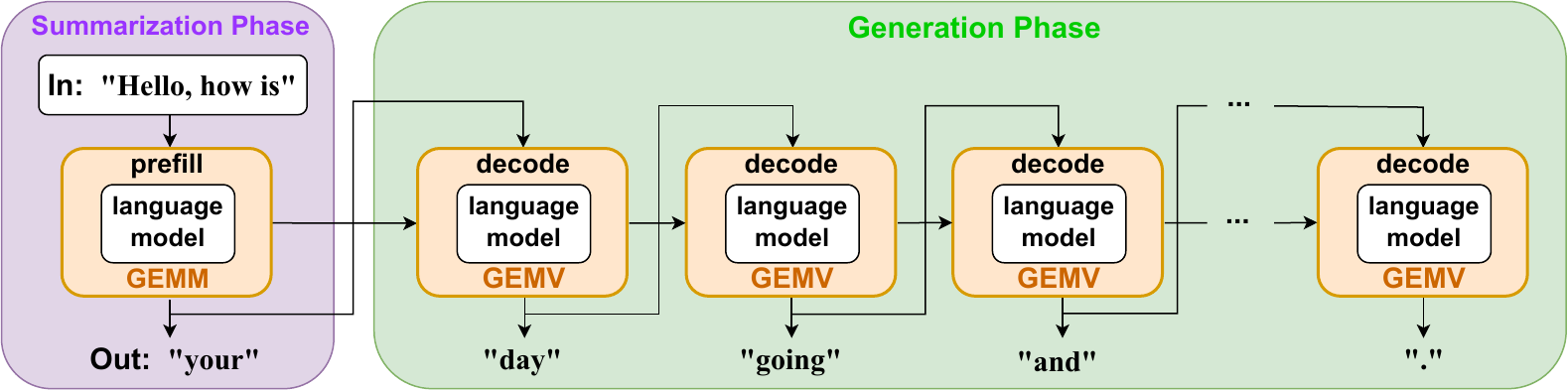}
    \caption{Illustration of LLM inference.} 
    \label{fig:LLM_infer}
\end{figure}

Recent efforts \cite{rouhani2023microscaling, wu2023zeroquant, wang2024outliertune, nair2025matryoshka} indicate that 6-bit quantization can serve as a superior compromise, offering further acceleration potential over 8-bit methods while better preserving model accuracy compared to 4-bit schemes. Consequently, 6-bit quantization is increasingly seen as a promising trade-off between inference efficiency and model accuracy.  However, mainstream GPU architectures currently lack native support for INT6 tensor operations, forcing implementations to emulate 6-bit precision using higher-precision units, and thus significantly limiting the full exploitation of 6-bit compression benefits.

Moreover, approximately 80\% of the computation and parameter access in LLMs is concentrated on general matrix multiplication (GEMM) and vector multiplication (GEMV) operations \cite{xia2023flash, zeng2025abq}. As shown in Figure \ref{fig:LLM_infer}, During autoregressive decoding, all GEMM operations effectively degenerate into GEMV operations for single-token generation. As such, the efficiency of GEMV computation and memory access critically determines overall LLM inference performance. To maximize the computational efficiency of GEMV and GEMM operations, recent studies \cite{yao2022zeroquant, xiao2023smoothquant, shao2024omniquant, lin2024duquant, zeng2025abq} have predominantly focused on full quantization of both weights and activations. Compared to weight-only quantization approaches \cite{frantar2022gptq, chee2023quip, lin2024awq, lee2024owq, tseng2024quip}, full quantization affords additional benefits, including further reductions in memory footprint, alleviation of memory bandwidth constraints, and more effective utilization of hardware computational capabilities. For instance, NVIDIA GPUs exhibit significantly higher peak throughput in low-precision modes (e.g., INT8, INT4) via Tensor Cores (TCs) relative to FP16 \cite{ampere_white}. However, in GEMV inference scenarios (particularly with small batch sizes), the computational efficiency is constrained by NVIDIA’s hardware architecture. Specifically, when batch sizes are below 8, TC units require padding operations to meet their fixed computational granularity, leading to two key issues: (1) underutilization of computational resources due to reduced effective workload; and (2) unnecessary overheads associated with padding operations. These factors collectively diminish hardware utilization efficiency and result in substantial performance degradation during GEMV computations.

\begin{table}[ht]
    \setlength{\tabcolsep}{1.3mm}
    \small
    \center
    \begin{tabular}{llccccc}
        \hline
        \#Bit       & Method        & 1-7B      & 1-13B     & 1-30B     & 2-7B      & 2-13B     \\ \hline
        FP16        & -             & 5.68      & 5.09      & 4.10      & 5.47      & 4.88      \\ \hline
        \multirow{3}{*}{\shortstack{W6A6}} 
                    & SmoothQuant   & 6.03      & 5.42      & 4.55      & 6.20      & 5.18      \\
                    & OmniQuant     & 5.96      & 5.28      & 4.38      & 5.87      & 5.14      \\
                    & I-LLM         & 5.84      & 5.23      & 4.32      & 5.68      & 5.10      \\ \hline
        \multirow{3}{*}{\shortstack{W4A4}} 
                    & SmoothQuant   & 25.25     & 40.05     & 192.40    & 83.12     & 35.88     \\
                    & OmniQuant     & 11.26     & 10.87     & 10.33     & 14.26     & 12.30     \\
                    & I-LLM         & 9.10      & 7.99      & 7.24      & 10.44     & 9.76      \\
        \hline
    \end{tabular}
    \caption{Perplexity results of various quantization methods applied to the LLaMA family models with 6-bit and 4-bit weight-activation precision, evaluated on the WikiText2 dataset.}
    \label{tab:comparison_between_6bit_and_4bit}
\end{table}

In this work, we introduce FlexQ, a novel and efficient post-training INT6 quantization framework tailored for LLM inference. To enhance model accuracy under 6-bit quantization, we implement a fine-grained group quantization strategy for both weights and activations, exploiting local data characteristics to reduce quantization errors. Further, through an analysis of the model's linear layers, we observe that different layers exhibit varying sensitivities to 6-bit quantization. We leverage uniform weight compression across all layers and selectively preserve 8-bit activations in the most quantization-sensitive layers based on layer-wise sensitivity analysis. This approach ensures minimal accuracy loss while optimizing resource utilization. Additionally, we develop a specialized software engine that performs W6Ax inference based on BTC equivalents. This engine effectively eliminates redundant computations caused by padding in small-batch scenarios, thereby fully unlocking the performance potential of quantized models operating with INT6 mixed precision.

In summary, we make the following contributions.
\begin{itemize} 
\item We propose a fine-grained group quantization method based on 6-bit precision, exploiting local data features to strike an optimal balance between inference accuracy and efficiency.  Our quantization method employs a mixed-precision strategy by allocating higher bitwidths, such as INT8, to quantization-sensitive layers, effectively alleviating their impact on overall model accuracy. 
\item We develop a specialized GPU kernel that enables fast matrix multiplication for W6A6 and W6A8 representations based on BTC equivalents to support fully quantized inference under INT6 mixed precision. Through careful data layout design and computational scheduling, our kernel addresses the low utilization issues typical in small batch inference. 
\item  We perform a comprehensive evaluation on the LLaMA family and demonstrate that FlexQ achieves near-FP16 perplexity (with an increase of at most 0.1) on WikiText2, indicating negligible accuracy degradation. Our kernel significantly accelerates inference, delivering a 1.39× speedup over existing approaches like ABQ-LLM. Furthermore, end-to-end evaluations reveal that FlexQ surpasses state-of-the-art methods like SmoothQuant, delivering 1.33× faster inference and 1.21× greater memory savings. 
\end{itemize}

%-------------------------------------------------------------------------------
\section{Background and Related Work}
%-------------------------------------------------------------------------------

\subsection{Quantization of Large Language Models}

Although LLMs have demonstrated remarkable performance, their massive size poses significant deployment challenges. Model quantization is a widely adopted technique to address these issues by producing more compact model representations. The primary focus of quantization is on the weights of linear layers (i.e., matrix multiplication), which account for approximately 99\% of the total model weights \cite{lin2024qserve}. Activations can also be quantized during inference to further improve efficiency \cite{nagel2021white}. Quantization schemes can be categorized into symmetric and asymmetric approaches based on whether a zero point is employed to accommodate data distribution asymmetry \cite{dettmers2022gpt3, zhao2024atom, liu2025llmeasyquant}. Asymmetric quantization incorporates zero-point offsets to better capture the dynamic range of the data, thereby offering superior numerical precision. In contrast, symmetric quantization eliminates the zero-point, simplifying computations and making it more hardware-efficient. 
In this paper, we adopt symmetric quantization for both weights and activations. We denote the precision configuration using the notation ``\textbf{WxAy}", where \textbf{x/y} indicates the bit-width for weights and activations, respectively.

\subsection{NVIDIA GPU Architecture and Tensor Core}
\textbf{GPU Execution Hierarchy.}
NVIDIA GPUs are optimized for large-scale parallel computing, comprising multiple Streaming Multiprocessors (SMs), each integrating CUDA cores, Tensor Cores, and other specialized functional units. The execution model employs a hierarchical thread organization: individual threads serve as the fundamental execution units, with 32 threads forming a warp that executes instructions synchronously in lockstep \cite{cuda_programming_guide}. Multiple warps constitute a thread block, which shares on-chip memory resources. When launching a kernel, the configuration of the number of thread blocks and threads per block must be specified. Each thread block is assigned to a specific SM for execution, where threads within the same warp operate synchronously under the Single Instruction, Multiple Threads (SIMT) execution model. 

\textbf{Memory Hierarchy.}
The GPU memory hierarchy is a multi-level structure encompassing global memory, shared memory, and registers. Global memory, typically DRAM, provides the largest capacity but exhibits the highest latency. Its accesses are accelerated through an L2 cache and are globally visible to all SMs. Each SM is equipped with a dedicated L1 cache and a configurable shared memory region, which is accessible to all threads within a thread block. Shared memory adopts a 32-bank architecture, where each bank manages a 4-byte segment of contiguous data. During shared memory accesses, if multiple requests within a transaction target the same bank (referred to as bank conflicts), the transaction is split into multiple operations. These conflicts can substantially degrade the effective bandwidth and throughput of concurrent memory accesses, thereby impacting overall computational efficiency.

\textbf{Tensor Core.}
Tensor Cores are specialized units designed to accelerate matrix-matrix multiplication operations fundamental to neural network computation, providing significantly higher throughput and support for varied precision formats compared to conventional CUDA cores \cite{votal_white}. The dense TC, introduced in the Volta architecture, is optimized for general matrix multiplication and can deliver peak TFLOPS up to 6× higher than FP16 operations on CUDA cores \cite{votal_white}. To support more quantized neural networks, subsequent architectures like Turing \cite{jia2019dissecting} expand TC capabilities to include additional precision formats such as INT1, INT4, and INT8. Specifically, in the Turing architecture, TCs only support \texttt{XOR} logic operations for INT1. The Ampere architecture \cite{ampere_white} further extends this support by incorporating \texttt{AND} logic operations for INT1. Our work focuses on binary tensor cores (BTCs) with INT1, which utilize a unique binary computation paradigm. Compared to INT4 and INT8 TCs, BTCs achieve 4× and 8× higher peak computational throughput, respectively \cite{ampere_white}.

\subsection{Related Work}
\textbf{Weight-only Quantization.}
Weight-only quantization approximates weight matrices with low-bit representations to reduce computational complexity and memory footprint. GPTQ \cite{frantar2022gptq} utilizes 4-bit integers combined with Hessian-based error compensation to minimize quantization errors in LLMs. AWQ \cite{lin2024awq} and OWQ \cite{lee2024owq} emphasize the importance of quantizing weights associated with higher-magnitude activations, significantly improving quantized model performance. SqueezeLLM \cite{kim2024squeezellm} alleviates the performance degradation caused by quantization by preserving outliers and sensitive weight values within sparse matrices. QuIP \cite{chee2023quip} and QuIP\# \cite{tseng2024quip} achieve 2-bit quantization by employing learnable codebooks and additional fine-tuning, often coupled with vector quantization of weights. QuaRot \cite{ashkboos2024quarot} implements INT4 quantization using Hadamard transformations to effectively handle outliers in linear and attention layers. Recently, Quant-LLM \cite{xia2024quant} proposes the first GPU kernel designed specifically for FP6 weight quantization. In contrast, our work applies INT6 quantization to weights.

\textbf{Weight-activation Quantization.}
In contrast to weight-only schemes, W/A quantization targets both weights and activations to further enhance compression and efficiency. ZeroQuant \cite{yao2022zeroquant} employs fine-grained quantization with variable precisions for weights and activations, optimizing overall performance. SmoothQuant \cite{xiao2023smoothquant} introduces a transformation that shifts the quantization difficulty from activations to weights, enabling practical 8-bit (W8A8) quantization. OmniQuant \cite{shao2024omniquant} and Atom \cite{zhao2024atom} explore aggressive W4A4 and mixed-precision W4A8 schemes, respectively. QServe \cite{lin2024qserve} adopts a progressive W4A8KV4 quantization approach combined with smooth processing for attention layers, effectively mitigating accuracy degradation caused by 4-bit quantization. I-LLM \cite{hu2024llm} achieves fully integer-based inference through smooth block reconstruction and complete reliance on integer operators. SpinQuant \cite{liu2024spinquant} and OSTQuant \cite{hu2025ostquant} introduce learnable rotations embedded within the network to maximize quantization accuracy. Meanwhile, COMET \cite{liu2025comet} facilitates practical deployment of W4A4KV4 models by leveraging channel permutation techniques and auxiliary optimization strategies. DuQuant \cite{lin2024duquant} applies rotation and permutation transformations to better mitigate outliers. ABQ-LLM \cite{zeng2025abq} decomposes low-bit quantized weights and activations into binary matrix representations to enable efficient arbitrary-precision inference. Our approach differs from these methods by exploring adaptive fine-grained group quantization schemes for both weights and activations.

\section{Opportunities and Challenges}
\label{sec:moti}
In this section, we first analyze the potential benefits of 6-bit quantization for efficient LLM deployment by comparing it against existing methods (i.e., 8-bit and 4-bit quantization). Then, we identify the key challenges in realizing high-performance, 6-bit quantized LLM inference on modern mainstream GPUs. These insights serve as the primary motivation for our proposed system design.

\subsection{6-bit Quantization: a Superior Trade-Off Between Model Quality and Inference Efficiency}
The deployment of LLMs is bottlenecked by their massive memory footprints and computational demands. While quantization offers a viable solution to mitigate this bottleneck, existing methods still face a dilemma: 4-bit quantization \cite{frantar2022gptq,lin2024awq,liu2025comet} sacrifices model accuracy for efficiency, whereas 8-bit quantization \cite{dettmers2022gpt3,yao2022zeroquant,xiao2023smoothquant} preserves accuracy at higher resource costs and lower inference efficiency. Recent algorithmic advances \cite{rouhani2023microscaling,wu2023zeroquant,wang2024outliertune,nair2025matryoshka} demonstrate that 6-bit quantization can strike a superior balance between inference efficiency and model quality, highlighting promising opportunities for LLM serving beyond traditional 8-bit and 4-bit methods.

Compared to 8-bit quantization, adopting more aggressive 6-bit schemes can further reduce deployment costs without significant accuracy loss. Specifically, the size of LLM weights can be compressed to approximately 2.7× smaller than the FP16 baseline, substantially decreasing GPU memory requirements and thereby reducing the number of GPUs needed for deployment. Moreover, 6-bit quantization enables more efficient inference acceleration (particularly for the memory-bound token-generation phase) by reducing GPU DRAM accesses.  As illustrated in Figure \ref{fig:kernel_benchmark} (Section \ref{sec:Kernel_Performance}), the linear layer execution within the LLaMA-70B model \cite{touvron2023llama} is consistently faster (up to 1.39×) using our proposed method (FlexQ-W6Ax) compared to the SoTA baseline (ABQ-LLM). Since linear layers constitute the majoritity of computational load in LLMs, this speedup can directly translate to improved end-to-end inference performance, as demonstrated in Section \ref{sec:E2E_Evaluation}.

While 4-bit quantization reduces memory footprint and DRAM access even further, it inherently compromises model quality. Conversely, 6-bit quantization affords near-lossless compression. As shown in Table \ref{tab:comparison_between_6bit_and_4bit}, INT6 methods exhibit robust and consistent performance across different model scales, such as 1B, 13B, and 30B LLaMA models while all INT4 quantization methods suffer from substantial accuracy loss.  Moreover, INT4 methods rely heavily on fine-grained quantization techniques to maintain performance \cite{frantar2022gptq,zhao2024atom,lin2024qserve,liu2025comet}, whereas INT6 methods can still perform effectively even with coarse-grained quantization.

In summary, 6-bit quantization provides a practical and promising alternative for democratizing LLM deployment by delivering substantial resource savings with minimal impact on model quality.

\subsection{Design Challenges}
\label{moti:challenges}
While INT6 quantization offers promising opportunities for further model compression and hardware efficiency, realizing its full potential presents three major challenges. First, the reduced quantization levels (from 256 in INT8 to 64 in INT6) complicate accuracy preservation, primarily due to activation outliers and the need for layer-adaptive strategies to handle the heterogeneity of network sensitivities. Second, the lack of dedicated hardware primitives for INT6 operations means that implementing INT6 on existing hardware often relies on software-level data packing or emulation strategies using higher-precision arithmetic units, which deliver suboptimal computational throughput.

Third, modern GPU memory systems do not naturally support irregular bit-widths (non-powers of two), since the minimal access size in global or shared memory is typically 8 or 32 bits per thread, and memory accesses must be properly aligned. The complex data layout requirements of tensor cores further exacerbate  the challenge of efficient implementation for irregular bit-widths.

\begin{figure}[h]
    \centering
    \includegraphics[width=\linewidth]{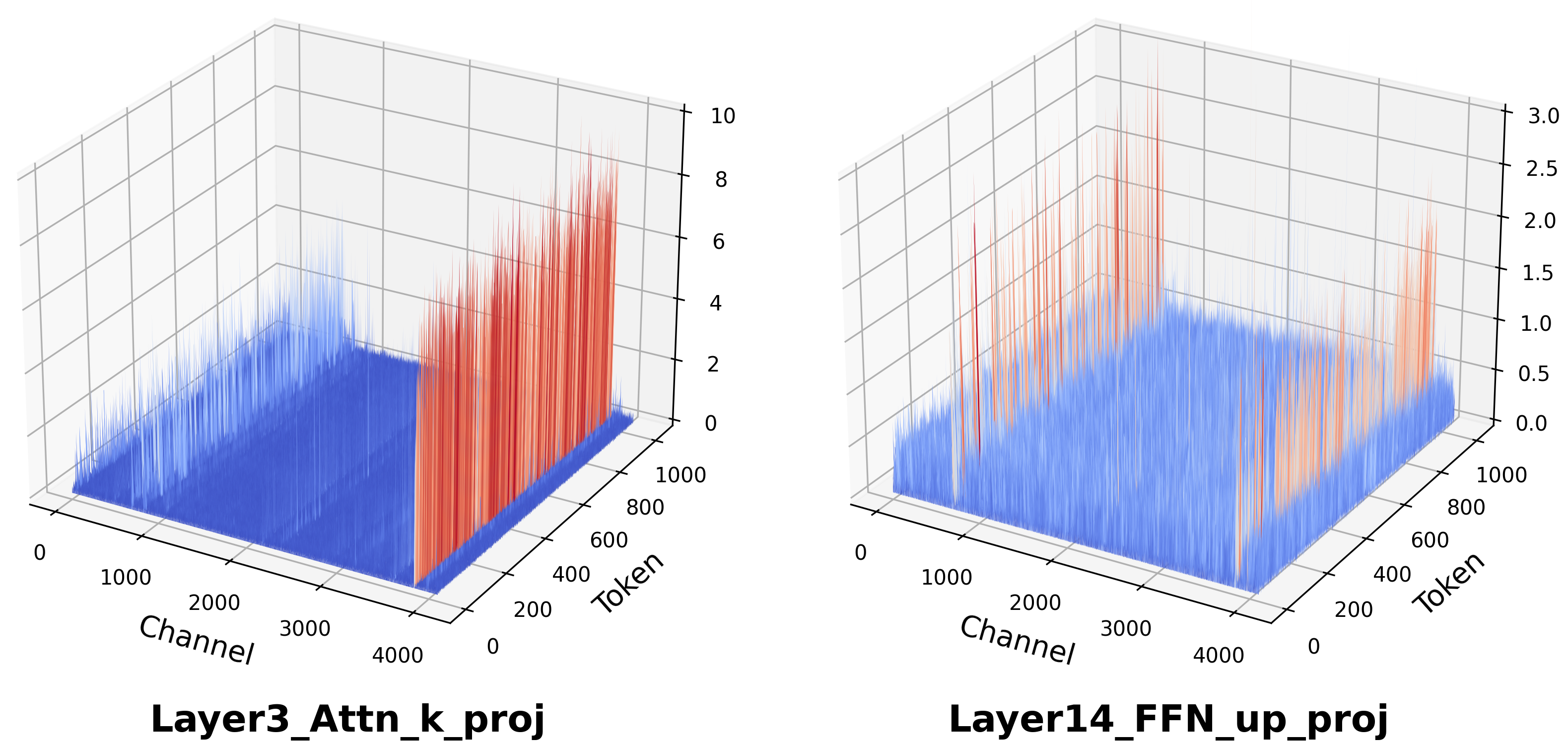}
    \caption{The input activation distribution in the LLaMA-7B model displays pronounced outliers across all token sequences, particularly within the k\_proj of layer 3 and the up\_proj of layer 14.}
    \label{fig:activation_distribution}
\end{figure}

\textbf{Challenge\_1: Model Quality Preservation in 6-bit Full Quantization.}  INT6 provides only 64 quantization levels compared to 256 in INT8, significantly constraining its representational capacity. This limited range heightens the susceptibility to quantization errors, particularly for high-variance parameters and outlier.  As illustrated in Figure \ref{fig:activation_distribution}, activations often contain a few channels with significantly large magnitudes, resulting in a wide dynamic range \cite{xiao2023smoothquant, zhao2024atom}. These outliers can significantly increase quantization error because their large values tend to dominate the quantization process, leading to degraded model accuracy. INT8 quantization commonly employs outlier mitigation techniques such as outlier clipping or mixed-precision schemes, which help preserve overall accuracy. However, at INT6, these methods become considerably more sensitive. Clipping can cause substantial information loss if not carefully tuned, while mixed-precision approaches may add complexity beyond current hardware support. Achieving a stable trade-off between effective outlier handling and minimizing quantization error is therefore more challenging.

Moreover, different layers within neural networks exhibit diverse distributions and sensitivities to quantization \cite{sun2024massive, yang2024mitigating, zeng2025abq}. While INT8 models can often utilize uniform layer-wise quantization settings with minimal accuracy loss, the narrower dynamic range of INT6 demands more adaptive layer-specific schemes. Such per-layer quantization strategies are critical for maintaining model accuracy in low-bit regimes.

\textbf{Challenge\_2: Hardware Support and Deployment Limitations.} 
Despite the architectural advantages of 6-bit weight-activation quantization, its deployment in practical scenarios faces significant hurdles. Currently, mainstream NVIDIA GPUs predominantly support data types like FP16, INT8, and INT4 \cite{hanindhito2024accelerating}, but lack native support for 6-bit tensor core MMA operations. This absence requires software-based data packing or emulation strategies that leverage higher-precision arithmetic units, resulting in reduced computational throughput and diminishing the efficiency benefits. For instance, the NVIDIA A100 GPU offers a peak throughput of 1248 TOPS at INT4, whereas at INT8, it provides only 624 TOPS. Moreover, mixed-precision schemes such as W4A8 \cite{lin2024qserve} require dequantization of low-bit parameters to higher-precision formats before performing computation. This process introduces additional computational overhead, diminishing the overall efficiency advantage of lower-bit quantization.

Although NVIDIA's recent Blackwell GPU architecture \cite{blackwell_white} introduces native FP6 tensor cores optimized for 6-bit quantization, its industry adoption remains limited. The transition to new hardware architectures involves prolonged deployment cycles, with most existing infrastructure still based on conventional GPU designs. Moreover, the high total cost of ownership associated with Blackwell devices hinder large-scale deployment, reducing economic viability for widespread adoption. 

This disconnect between the promise of 6-bit quantization and current hardware capabilities poses a critical challenge: enabling efficient and hardware-supported inference at this bit-width on widely used GPU platforms without sacrificing acceleration benefits.  Addressing these hardware and deployment constraints is imperative to unlocking the full potential of 6-bit quantization in practical and industrial applications.

\textbf{Challenge\_3: Hardware-Unfriendly Memory Access for 6-bit Quantization.} While 6-bit full quantization schemes substantially compress models and enhance computational throughput, their irregular bit-width introduces significant compatibility and efficiency challenges within modern GPU memory hierarchies. Contemporary GPU memory systems  (comprising global memory, shared memory, and register files) are optimized around power-of-two bit-width units, primarily aligning data transfers with 8-bit, 16-bit, or 32-bit boundaries to maximize memory bandwidth utilization and minimize latencies \cite{cuda_programming_guide}. The non-standard 6-bit quantization disrupts this alignment, leading to two primary issues.

First, efficient memory access relies on aligned data transactions. Since 6 bits do not align with the conventional 8-bit or 32-bit boundaries, accessing these weights involves irregular, cross-boundary reads. Such unaligned access patterns degrade performance by causing a single logical data request to be decomposed into multiple disjointed transactions. This fragmentation increases memory access latency and diminishes the benefits expected from low-bit quantization.

Second, due to the irregular data packing, loading weights in 6-bit format results in substantial redundant data transfers. For example, in shared memory, loading two 6-bit weights ideally should consume 12 bits. However, because shared memory banks operate on 32-bit transaction units, part of the bandwidth is underutilized. Specifically, a 32-bit transaction contains only 12 useful bits for the two weights, wasting approximately 20 bits per transaction and reducing bandwidth utilization to around 37.5\%. The problem exacerbates when weights span across multiple memory banks or boundaries. For instance, a thread requiring weights split between banks must perform additional transactions, further amplifying bandwidth waste and latency and reducing utilization to as low as 18.75\% in worst-case scenarios.

\begin{figure}[h]
    \centering
    \includegraphics[width=\linewidth]{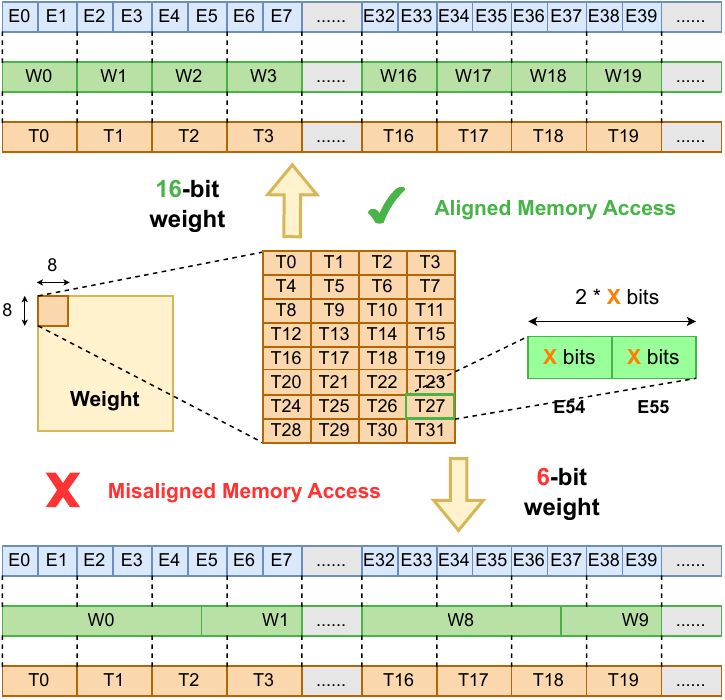}
    \caption{Memory access analysis of data loading from shared memory to registers on a per-thread basis. Here, E represents a weight element, W denotes a 32-bit word, and T indicates a thread.}
    \label{fig:memory_access}
\end{figure}

These inefficiencies are not confined solely to shared memory but also extend to global memory accesses and register loads, manifesting as fundamental incompatibilities between 6-bit data schemes and existing memory architectures. This disconnect hampers the ability of current GPU hardware to fully exploit the low-bit-width benefits of 6-bit quantization, presenting a critical obstacle to its practical deployment for large-scale LLM inference. A detailed comparison of memory access behaviors (as shown in Figure \ref{fig:memory_access}) illustrates this challenge. Under the standard \textit{mma.m16n8k8} computation mode, FP16 weights align seamlessly with 32-bit memory boundaries, enabling each thread to load two elements per transaction efficiently and thus achieving nearly 100\% memory bandwidth utilization. Conversely, 6-bit weights, due to their irregular packing, force each thread to handle fragmented data loads, severely impairing bandwidth efficiency and increasing latencies. Addressing these issues requires novel hardware-aware data packing and memory access strategies tailored specifically for non-standard bit-widths like 6 bits.

% \section{Method}
\section{FlexQ: An Accuracy and Efficiency Co-Designed  Quantization System}
Our key observation is that 6-bit quantization offers both significant opportunities and formidable challenges for efficient LLM serving (as detailed in Section~\ref{sec:moti}). Its high compression ratio to significantly reduce memory bandwidth can substantially accelerate inference and lower deployment costs. However, the irregular, low-bit width not only heightens the risk of accuracy degradation but also introduces serious compatibility issues with modern GPU memory systems, resulting in inefficient memory accesses, underutilized bandwidth, and increased latency.

To capitalize on the benefits of 6-bit quantization while addressing its challenges, FlexQ employs delicated quantization strategies and a suite of system-level optimization techniques. These include: (1) a adaptive fine-grained group quantization scheme for both weights and activations to preserve model accuracy (Section~\ref{sec:quant_method}), (2) a delicated software engine to support quantized inference of INT6 based on BTC equivalents (Section~\ref{sec:bit_operation_template}), and (3) a highly optimized GPU kernel designed to enable efficient matrix multiplication for W6A6 and W6A8 representations (Section~\ref{sec:kernel_design}). Collectively, these innovations  mitigate the memory access inefficiencies associated with 6-bit scheme effectively, unlocking its full potential for high-performance, resource-efficient LLM serving without compromising model quality.

\subsection{Quantization Scheme}
\label{sec:quant_method}
In this section, we provide a systematic overview of our proposed fine-grained group quantization approach, which enhances the capability to preserve model accuracy under W6A6 quantization. We then introduce a high-precision activation quantization strategy tailored for layers that are particularly sensitive to quantization-induced errors, dynamically adjusting the quantization precision based on layer-specific sensitivity. Finally, we demonstrate the integration of this comprehensive quantization scheme into the end-to-end inference pipeline of the LLaMA model, thereby showcasing the practical viability of FlexQ.

\subsubsection{Weight \& Activation Fine-grained Group Quantization}
The choice of quantization granularity for weights and activations plays a crucial role in balancing model accuracy and computational efficiency. Conventional coarse-grained quantization schemes, such as per-token or per-channel quantization, often lead to non-ignorable accuracy degradation, even at 6-bit precision, primarily due to their inability to accurately capture the dynamic range within entire channels with a single scaling factor. 

To address this limitation, we adopt a fine-grained group quantization strategy \cite{lin2024awq,zhao2024atom,frantar2025marlin}, which has been shown to better preserve model quality. Specifically, FlexQ applies a systematic and more granular grouping of elements within each channel. Consecutive elements are partitioned into groups of size n, where each group shares a common scaling factor. For example, with a group size of 128, every sequence of 128 consecutive elements in a channel constitutes a subgroup for quantization, allowing the local data distribution to be more accurately modeled. This approach effectively maintains model accuracy by capturing local variations within each group and hence reducing quantization error substantially.

However, this improved model accuracy comes with a trade-off: the increased overhead due to the necessity of applying individual dequantization operations per group. To mitigate this overhead, we integrate the dequantization process directly into the GEMM kernel, enabling efficient in-kernel dequantization without additional data movement overhead. The detailed implementation and optimization strategies are elaborated in Section~\ref{sec:tc_dequant}.

\subsubsection{Selective High-Precision Activation Quantization for Sensitive Network Layers}
While the 6-bit fine-grained group quantization strategy effectively preserves overall model accuracy, our in-depth analysis of large model architectures reveals that, under 6-bit precision, different network layers exhibit varying sensitivities to quantization. Notably, layers such as qkv\_proj within attention modules, as well as gate\_proj and down\_proj in feed-forward network (FFN) modules, demonstrate significant disparities in their resilience to quantization effects. This observation motivates the exploration of higher-precision quantization schemes specifically for these quantization-sensitive layers.

\begin{figure}[ht]
    \centering
    \includegraphics[width=\linewidth]{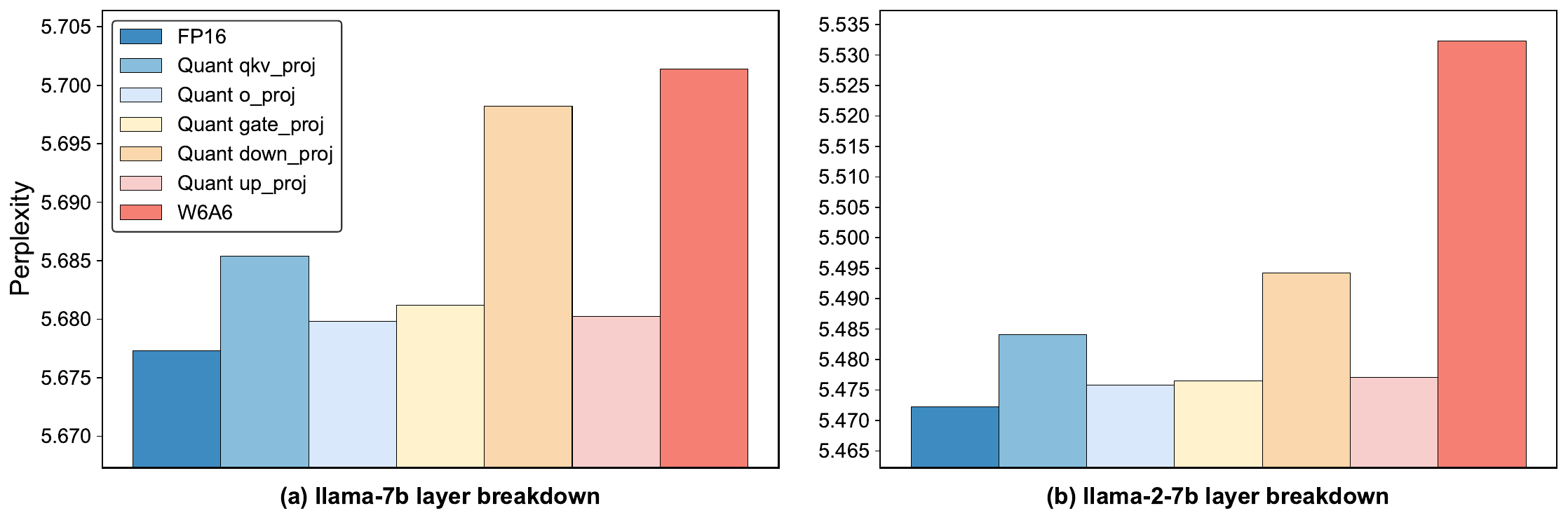}
    % \caption{Perplexity results (lower is better) for quantizing various linear layers in the LLaMA-7B and LLaMA-2-7B models on the WikiText2 dataset.}
    \caption{Perplexity results (lower is better) for the LLaMA-7B and LLaMA-2-7B models, evaluating various linear layers on the WikiText2 dataset.}
    \label{fig:linear_layer_breakdown}
\end{figure}

To identify such sensitive components, we analyze both LLaMA-7B and LLaMA-2-7B models. We apply fine-grained symmetric quantization to weights and activations across different linear layers, evaluating impact via perplexity metrics. As shown in Figure \ref{fig:linear_layer_breakdown}, quantizing layers such as o\_proj, gate\_proj, and up\_proj in the respective attention and FFN modules results in minimal accuracy degradation. Conversely, quantizing qkv\_proj and down\_proj leads to more pronounced performance declines. Among these, the down\_proj layer exhibits the highest sensitivity, indicating it is the most critical to maintain higher precision.

Previous studies \cite{yang2024mitigating, lin2024duquant} show that in GLU-based LLMs (e.g., LLaMA, Mistral, Mixtral, Gemma), activation outliers frequently occur at the input to the FFN down\_proj layer. The large dynamic range of these outliers significantly amplifies quantization errors. Based on this insight, we propose to retain higher precision for the input activations of the down\_proj layer. To balance efficiency and accuracy, we adopt a differentiated quantization strategy: quantizing these activations to 8 bits while applying 6-bit quantization to other linear layers. This approach effectively mitigates accuracy loss in sensitive layers through such selective high-precision activation quantization. 

It is important to note that the quantization strategy employed in FlexQ is calibration-free and does not depend on any external dataset. While calibration-based methods such as AWQ \cite{lin2024awq}, GPTQ \cite{frantar2022gptq}, and ABQ-LLM \cite{zeng2025abq} can effectively preserve accuracy, they face notable limitations: (1) dependence on calibration data, which may introduce distribution bias and degrade model quality, (2) reliance on high-performance hardware for calibration, limiting deployment on edge devices and lightweight scenarios. Our calibration-free approach offers a streamlined and resource-efficient solution, enabling the deployment of large-scale models without the complexities or hardware requirements associated with calibration procedures.

\subsubsection{FlexQ System Runtime}
\begin{figure}[ht]
    \centering
    \includegraphics[width=\linewidth]{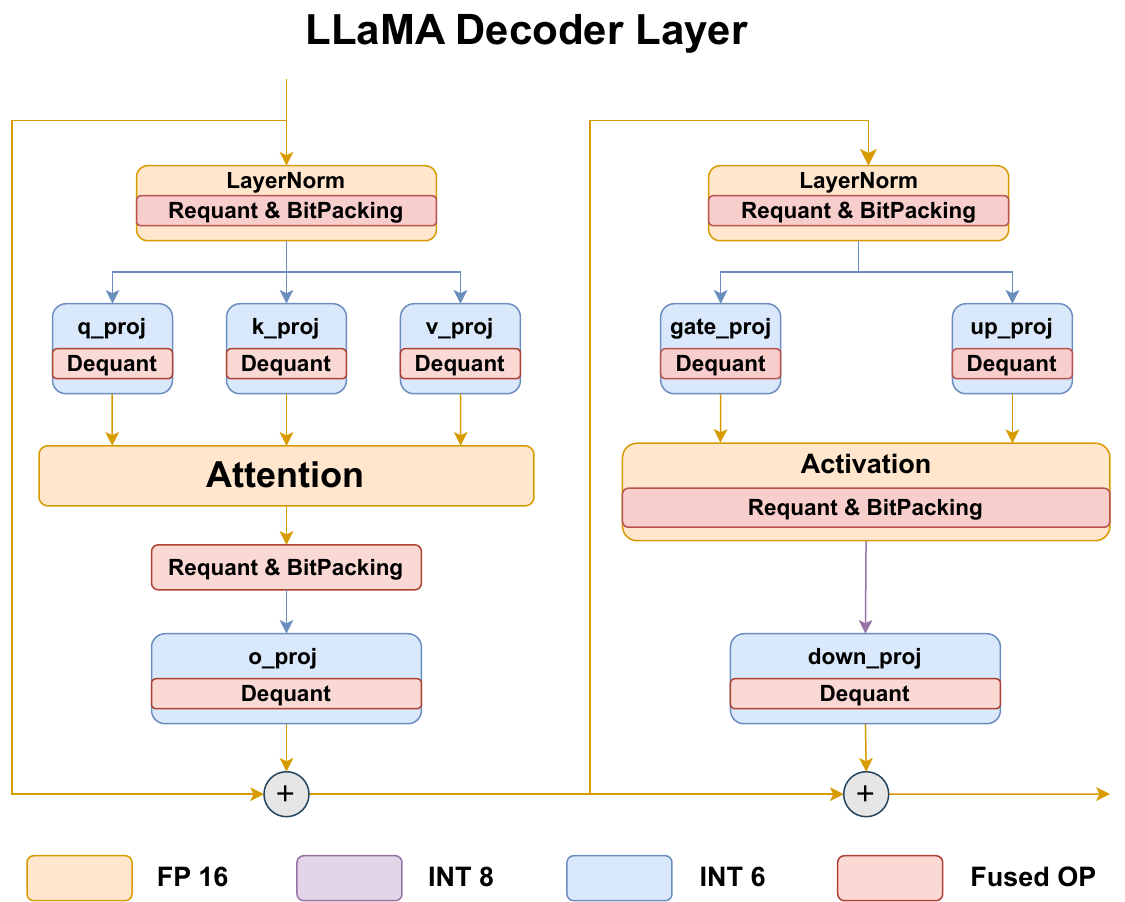}
    % \caption{Overview of the FlexQ workflow on LLaMA, where ReQuant and DeQuant denote online quantization and dequantization operations, respectively, and BitPacking represents the online Bit-level Data Packing operation for activations.}
    \caption{Overview of the FlexQ workflow for LLaMA models, where ReQuant and DeQuant denote online quantization and dequantization operations, respectively, and BitPacking represents the online Bit-level Data Packing operation for activations.}
    \label{fig:workflow_overview}
\end{figure}

To systematically elucidate the FlexQ runtime design, we utilize the LLaMA model as the benchmark architecture for demonstrating our implementation, as illustrated in Figure \ref{fig:workflow_overview}. At the quantization precision strategy level, we adopt a global 6-bit quantization scheme for all dense layer weights. Leveraging differentiated precision control for sensitive layers, the input activations of quantization-critical layers are maintained at 8-bit precision, while the remaining dense layers are uniformly quantized to 6 bits. All GEMM operations within these quantization-critical layers (i.e., down\_proj) are executed on specialized binary tensor cores operating in the W6A8 precision mode. By tightly integrating the dequantization process into the GEMM kernel, the output tensors of dense layers can be efficiently restored to FP16 precision immediately after matrix multiplication and accumulation, minimizing overhead. All quantized bit-level data representations are managed via Bit-level Packing transformations, with implementation detailed in Section~\ref{sec:bit_packing}.

During inference, the static nature of weights allows for offline preprocessing, where weight quantization and BitPacking are performed prior to runtime. In contrast, activations, which are dynamically generated, require online processing, including real-time quantization, dequantization, and BitPacking. To optimize overall computational throughput, we employ kernel fusion techniques that fuse these operations within existing operators. Specifically, for qkv\_proj, gate\_proj, and up\_proj linear layers, activation quantization is fused into their preceding layer normalization kernel. For the down projection linear layer, the quantization operation is integrated into the activation function computation kernel. Notably, given the memory constraints within the self-attention module, we introduce a dedicated fused node tailored for activation quantization and BitPacking, positioned prior to the output projection linear layer. This fusion reduces memory bandwidth pressure and ensures seamless data processing within the constraints of GPU resources.

\subsection{Delicated Software Engine for INT6 Quantized Inference}
\label{sec:bit_operation_template}
While our INT6-based quantization strategy provides a superior trade-off balance between model accuracy and inference efficiency, a critical challenge is that current GPUs lack native support for INT6 tensor operations. To overcome this obstacle, we propose a specialized kernel that leverages bit-level decomposition techniques based on Binary Tensor Core (BTC) operations to efficiently support irregular bitwidths such as INT6. 

BTC, first introduced with NVIDIA's Turing architecture, has maintained backward compatibility across subsequent GPU generations. However, the inherently limited precision of W1A1 quantization, due to its difficulty in preserving model accuracy, has limited its practical deployment \cite{zeng2025abq}. Nonetheless, recent research on BTC \cite{feng2021apnn, wang2022qgtc, zeng2025abq} has proposed an alternative computational paradigm for scalar multiplication that decomposes operations at the bit level. This approach primarily involves three steps: (1) decomposing scalar values into their constituent bits, (2) performing bitwise multiplications for each bit position, and (3) aggregating the results through shifting and reduction operations.  By adopting this bit-level decomposition, our kernel effectively circumvents the lack of direct hardware support for W6A6 and W6A8 tensor operations, enabling efficient execution on existing GPU architectures via BTC-based computations.

To illustrate this methodology, consider two scalars: a 2-bit scalar $a$ and a 4-bit scalar $b$. The computational process begins with bit-level decomposition via weighted bit expansion, whereby binary numbers are expressed as weighted sums based on their positional weights (i.e., powers of 2). Specifically, scalars $a$ and $b$ can be mathematically represented as:

\begin{equation}
\begin{aligned}
a=\sum_{i=0}^{a\_bits-1}{a_i\cdot2^i}=a_1\cdot2^1+a_0\cdot2^0
\end{aligned}
\end{equation}

\begin{small} 
\begin{equation}
\begin{aligned}
b=\sum_{i=0}^{b\_bits-1}{b_i\cdot2^i}=b_3\cdot2^3+b_2\cdot2^2+b_1\cdot2^1+b_0\cdot2^0
\end{aligned}
\end{equation}
\end{small}

where $a\_bits$ and $b\_bits$ denote the bit-widths of scalars $a$ and $b$, respectively. Here, $a_i$ and $b_i$ are binary variables (0 or 1), representing the individual bits of the scalars after decomposition. Applying the distributive property of multiplication, the product of $a$ and $b$ expands as:

\begin{small} 
\begin{equation}
\scalebox{0.9}{$
\begin{aligned}
    a\cdot b&=\left(a_1\cdot2^1+a_0\cdot2^0\right)\cdot\left(b_3\cdot2^3+b_2\cdot2^2+b_1\cdot2^1+b_0\cdot2^0\right)\\
    &=a_1b_3\cdot2^4+\left(a_1b_2+a_0b_3\right)\cdot2^3+\left(a_1b_1+a_0b_2\right)\cdot2^2\\
    &\quad +\left(a_1b_0+a_0b_1\right)\cdot2^1+a_0b_0\cdot2^0
\end{aligned}
$}
\end{equation}
\end{small}

This formulation naturally generalizes to vector-vector and matrix-matrix multiplication across arbitrary bit-widths. Specifically, for matrix multiplication involving a p-bit weight matrix $W$ and a q-bit activation matrix $X$, the decomposition proceeds by extracting 1-bit weight matrices $W^{(s)}$ and activation matrices $X^{(t)}$, where $s \in \{0, 1, \ldots, p-1\}$ and $t \in \{0, 1, \ldots, q-1\}$. The key computation involves performing all binary matrix multiplications $p*q$ times, which can be mathematically expressed as:

\begin{equation}
\begin{aligned}
Y^{(s,t)}=bmma(W^{\left(s\right)},X^{(t)})
\end{aligned}
\end{equation}

where $bmma(\cdot)$ denotes a binary matrix multiplication operation accepting 1-bit inputs and producing 32-bit integer outputs. Finally, by aggregating these intermediate results weighted by their respective bit significance, the complete 32-bit integer output matrix $Y$ can be obtained using this equation:

\begin{equation}
\begin{aligned}
Y=\sum_{s=0}^{p-1}\sum_{t=0}^{q-1}{Y^{(s,t)}\ast2^{s+t}}
\end{aligned}
\end{equation}

Under our fine-grained group quantization strategy of FlexQ, the matrix multiplication formula must be accordingly adapted. Specifically, the data within each channel is divided into $g\_nums$ groups along the $K$ dimension, with each group containing $K / g\_nums$ consecutive elements. This grouping enables the binary matrix multiplication-accumulate ($bmma$) operations to be performed at a finer granularity corresponding to each group. Furthermore, we integrate dequantization directly into the GEMM computation, allowing the complete calculation to be expressed as:

\begin{equation}
\begin{aligned}
Y=\sum_{s=0}^{p-1}\sum_{t=0}^{q-1}\sum_{i=0}^{g\_nums-1}{S_{g_i}^W S_{g_i}^X\ast bmma\left(W_{g_i}^{(s)}X_{g_i}^{(t)}\right)}\ \label{con:bitlevelcal}
\end{aligned}
\end{equation}

where $S_{g_i}^W$ and $S_{g_i}^X$ denote the quantization scaling factors for the $i$-th group of the weight matrix $W_{g_i}$ and activation matrix $X_{g_i}$, respectively.

Building upon this computational framework, we implement the targeted fine-grained group quantization for W6Ax GEMM as a sequence of 1-bit matrix multiplication operations. Notably, in NVIDIA Turing architecture and subsequent GPU architectures, $bmma$ operations are natively supported at the hardware level. The specialized BTCs can efficiently execute these 1-bit matrix multiplications, achieving significantly higher throughput relative to traditional CUDA core implementations. This hardware acceleration facilitates our fine-grained group quantization scheme to attain superior performance in practical deployment scenarios.

\begin{figure*}[ht]
    \centering
    \includegraphics[width=\linewidth]{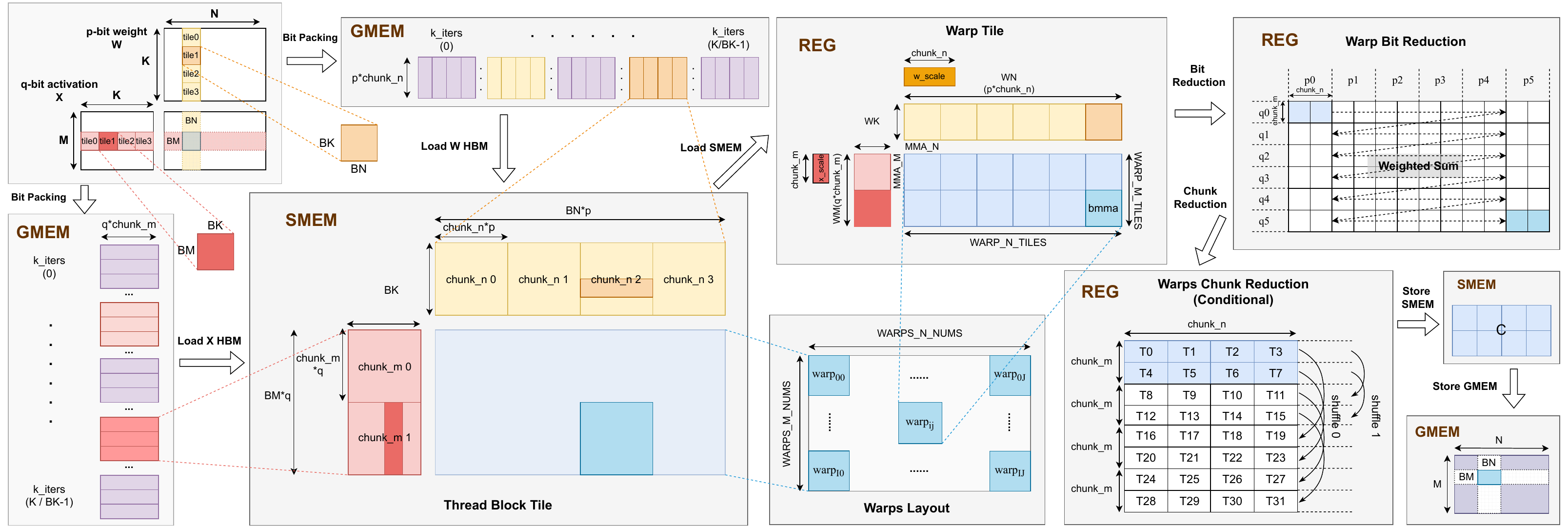}
    \caption{Overview of the FlexQ W6Ax kernel design.} 
    \label{fig:kernel_overview}
\end{figure*}

\subsection{FlexQ-W6Ax: Bit-Level Kernel Design}
\label{sec:kernel_design}
To address the critical challenge of inefficient memory access inherent to 6-bit quantization (Section~\ref{moti:challenges}), we develop a highly optimized mixed-precision FlexQ W6Ax kernel. As illustrated in Figure \ref{fig:kernel_overview}, our approach incorporates an innovative data layout with systematic optimizations to mitigate the memory bottleneck associated with 6-bit quantized models. Specifically, the kernel transforms the quantized tensor into a novel chunked bit-level data layout that ensures data access continuity and inherently prevents bank conflicts, thereby maximizing memory bandwidth utilization. We fuse the dequantization process into the computation phase of the BTC-based GEMM, reducing dequantization overhead without sacrificing accuracy. To compute bit-level weighted sums, we introduce an efficient warp-level reduction that leverages warp-level primitives for register-level data exchange to efficiently perform bitwise summation. Finally, a multi-stage software pipeline overlaps data loading with tensor core execution, further boosting computational throughput. Overall, this holistic approach significantly improves inference efficiency while maintaining the advantages of 6-bit quantization.

\subsubsection{Bit-level Data Packing}
\label{sec:bit_packing}
Building on the efficient implementation principles outlined in Section~\ref{sec:bit_operation_template} for the W6Ax compute engine, we details the preprocessing methodology for input weight and activation matrices. Specifically, we introduce a bit-level data packing strategy tailored to meet the architectural requirements of BTC units. This serves two primary objectives: (1) ensuring strict compliance of data formats with BTC specifications, and (2) optimizing in-memory storage patterns to maximize memory access throughput. By intelligently reorganizing data at the bit level, our approach effectively alleviates the memory access bottleneck associated with 6-bit quantization, thereby enabling faster LLM inference.

Specifically, when the group size is configured as 128 (the default setting in subsequent design descriptions), we employ the \texttt{m8n8k128} specification BTCs as the fundamental computing units, where the \texttt{MMA} dimensions are defined as \texttt{MMA\_M × MMA\_N × MMA\_K} (8 × 8 × 128). For an input activation matrix X (INT32 data type with shape [M, K]), the processing pipeline comprises three critical stages: First, bit-level decomposition extracts the binary representation of each element across different bit positions. Second, considering the characteristic that the M dimension is typically significantly smaller than the N and K dimensions \cite{xia2023flash,hong2023flashdecoding++,li2024evaluating} in LLM GEMM computations, we implement fine-grained tiling on both the M and K dimensions of the input matrix, partitioning it into multiple two-dimensional data chunks with dimensions  [chunk\_m, chunk\_k]. In practice, chunk\_m is set as min(M, MMA\_M), while chunk\_k is fixed at MMA\_K. This tiling strategy ensures that each BTC \texttt{MMA} operation on a chunk directly produces the corresponding single-bit computation result, establishing a foundation for subsequent warp-level reduction optimizations (as detailed in Section~\ref{sec:warp_reduction}). Finally, the partitioned data undergoes reorganization, transforming the original [M, K, q] shape of the activation matrix X into a [K/chunk\_k, M/chunk\_m, q, chunk\_m, chunk\_k] bit-level packed form. The weight matrix W is processed identically to preserve dimensional consistency. To facilitate efficient GPU execution, the implementation utilizes the \texttt{\_\_ballot\_sync} instruction to perform intra-warp bit-level data exchange and combination, ensuring continuous thread memory access patterns. This strategy enables memory transaction coalescing at the global memory level, maximizing data loading and computation throughput.

\subsubsection{Maximizing Loading and Storing Bandwidth}
\label{sec:max_memory_access}
For GPU kernels, efficiently loading weight and activation matrices into shared memory is critical for maximizing performance. A fundamental strategy is coalesced memory access. Since global memory resides in DRAM, achieving spatial locality during memory transactions is essential to optimize bandwidth utilization. For instance, when a matrix is stored in row-major order in global memory, threads should access data along rows sequentially to avoid cross-row strided accesses, which significantly degrade memory throughput. Additionally, employing vectorized load operations can further enhance efficiency by reducing the total number of instructions and increasing memory bandwidth utilization.

Specifically, for our GEMM kernel, given a weight matrix W and an activation matrix X in bit-level packed form for computation within a single thread block, we aim to optimize the memory access pattern to enable coalescing. Each thread performs load operations with a granularity of 16 bytes (128 bits). Consequently, in a single transaction, all threads (totaling threads\_num) collectively transfer 16*threads\_num bytes, equivalent to threads\_num INT4 elements. By applying bit-level data packing to both W and X, the data can be aligned according to structures such as p*chunk\_n*chunk\_k or q*chunk\_m*chunk\_k. This alignment ensures that memory accesses are contiguous across threads, thus facilitating efficient coalesced memory transactions.

In the data write-back phase for the output matrix C (size BM × BN), we adopt a similar memory access strategy as described above. Specifically, each thread within a block storing a segment of C writes data with a granularity of 16 bytes (128 bits) per operation. Leveraging this strategy, the overall write-back can be performed using BM*BN/(16/4) threads per block, ensuring coalesced global memory transactions.

Additionally, we observe suboptimal utilization of the L2 cache during GEMV or GEMM computations with very small batch sizes. A primary factor is the limited reuse of the weight matrix W: due to the small M dimension, each data block of W is accessed only once per kernel invocation. To prevent the weight matrix W from occupying precious L2 cache space unnecessarily and causing cache pollution, we employ the \texttt{evict\_first} cache eviction policy. By tagging the weight data with \texttt{evict\_first} priority hints, we designate it as the preferred candidate for eviction among eligible cache lines, effectively preventing it from polluting the cache with infrequently reused data. In contrast, activation matrices X, which generally exhibit higher reuse rates, do not require such eviction prioritization. This cache management strategy, therefore, can be extended to general LLM GEMM workloads, balancing cache occupancy and data reuse to optimize overall performance.

\subsubsection{Conflict-Free Shared Memory Layout}
\label{sec:shared_memory_layout}
Shared memory, as a critical on-chip resource, offers significantly lower access latency compared to global memory, thereby enhancing data communication efficiency within thread blocks. However, due to its banked architecture, typically comprising 32 banks, the data layout in shared memory directly influences access performance. Improper data arrangements can induce bank conflicts, leading to degraded memory throughput. Hence, designing an optimal shared memory layout is essential for maximizing GEMM computational performance.

\begin{figure}[ht]
    \centering
    \includegraphics[width=\linewidth]{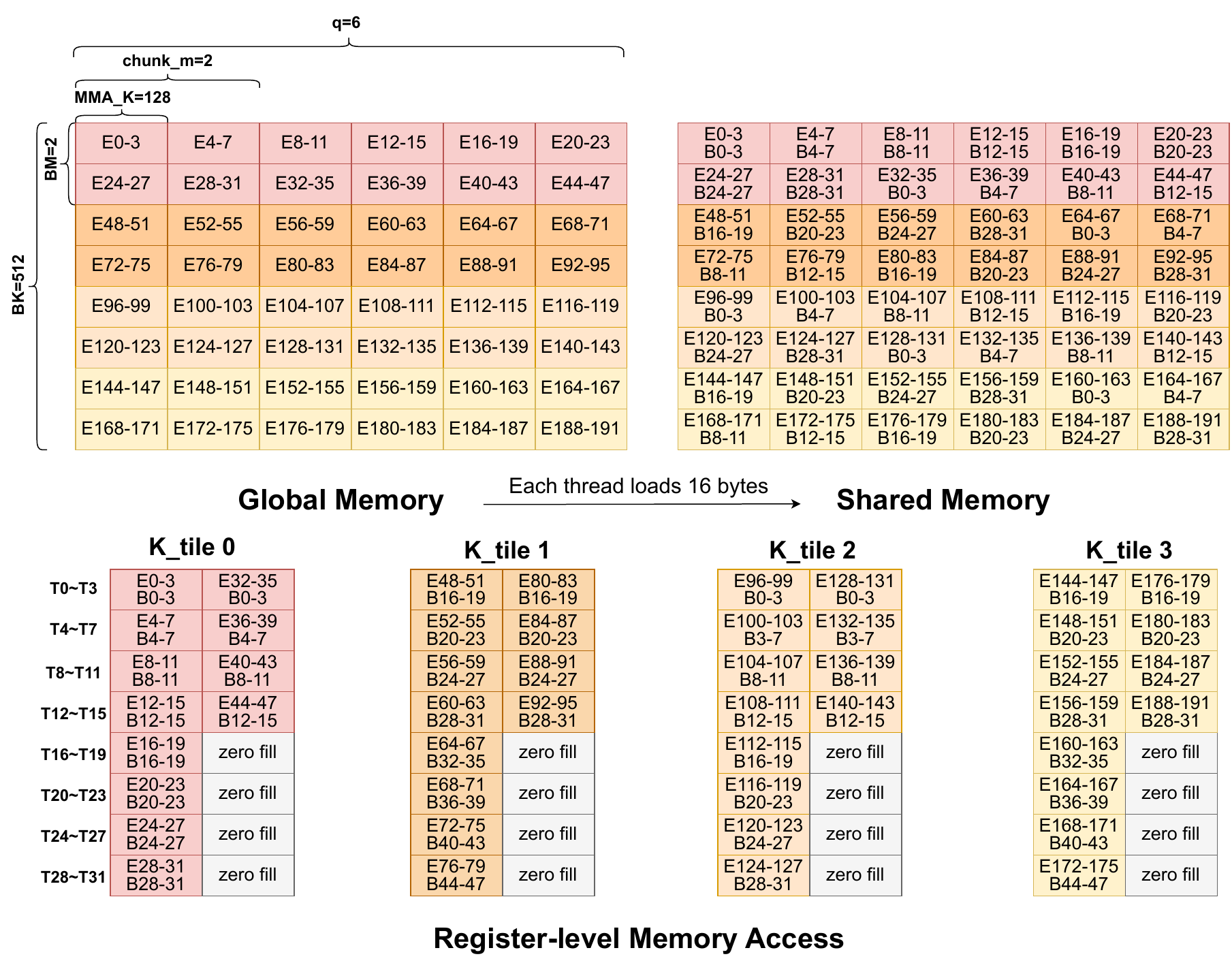}
    \caption{FlexQ's conflict-free shared memory layout design with zero fill indicating the redundant computation region, the bit-level design further minimizes the area of this region.}
    \label{fig:shared_memory}
\end{figure}

In our FlexQ kernel, we adopt a chunk-based data organization strategy to achieve conflict-free shared memory access. Specifically, the partial input activation matrix $X_{tile}$ and partial input weight matrix $W_{tile}$ are rearranged into layout shapes of [BK/\text{chunk\_k}, BM/\text{chunk\_m}, q, \text{chunk\_m}, \text{chunk\_k}] and [BK/\text{chunk\_k}, BN/\text{chunk\_n}, p, \text{chunk\_n}, \text{chunk\_k}], respectively. This design naturally prevents bank conflicts without requiring additional padding or swizzle operations \cite{lin2024qserve, zeng2025abq, du2025bitdecoding}.

As illustrated in Figure \ref{fig:shared_memory}, we demonstrate our layout's effectiveness using a 6-bit activation matrix with parameters BM = 2 and BK = 512. During global-to-shared memory loading, each thread loads 16 bytes, and the GPU decomposes the warp’s shared memory requests into a maximum of four transactions, each of 128 bytes. For instance, threads T0–T7 constitute one transaction, while T8–T15 form another, and so on. During shared memory-to-register transfer, each thread loads 4 bytes. Without redundant storage (i.e., no zero fill), these shared memory requests can be merged into a single maximum transaction. Notably, this layout prevents bank conflicts at all stages, ensuring continuous high-bandwidth memory access. Overall, this design guarantees coalesced memory access and optimal bandwidth utilization, thereby avoiding performance penalties associated with bank conflicts through precise dimensional alignment.

\subsubsection{Warp-Scheduled MMA and Fused Dequantization}
\label{sec:tc_dequant}
Once the warps load the necessary weight and activation matrices into registers, the Binary Tensor Core Matrix Multiply-Accumulate (BMMA) operation enables highly efficient bit-level multiplication by leveraging optimized instruction scheduling and hardware parallelism.

Taking the \texttt{mma.m8n8k128} shape  as an illustrative example, the compiler partitions each warp into 8 thread groups, each comprising four consecutive threads. During data loading, the four threads within each group cooperatively load segments of the weight and activation matrices along the MMA\_K dimension. Concurrently, the thread groups collaboratively handle data transfer along the [p, chunk\_n] and [q, chunk\_m] dimensions. Within a single iteration, each warp executes WARP\_M\_TILES*WARP\_N\_TILES BMMA operations, processing a data volume proportional to q*chunk\_m*p*chunk\_n*MMA\_K. The intermediate results from these operations are accumulated to produce the non-weighted GEMM output. 

FlexQ employs fine-grained group-wise quantization for both weights and activations. Each BMMA operation generates an intermediate result representing the multiplication of one weight group and one activation group, each associated with independently maintained quantization parameters (w\_scale and x\_scale). To eliminate the overhead associated with separate dequantization passes and additional memory transfers, FlexQ integrates dequantization directly into the accumulation stage of the intermediate results. Specifically, the quantization parameters w\_scale and x\_scale are stored as FP16 types, each thread loads the corresponding scales, which are then multiplied using the efficient half-precision parallel instruction \texttt{\_\_hmul2}. The resulting dequantized intermediate values are summed across threads to produce the final FP16 result. Furthermore, by extending computations across the bit dimension among thread groups, FlexQ reduces redundant calculations, particularly in scenarios where the batch size is less than 8, thereby further optimizing GPU hardware utilization.

\subsubsection{Warp-Level Reduction}
\label{sec:warp_reduction}
\begin{figure}[ht]
    \centering
    \includegraphics[width=\linewidth]{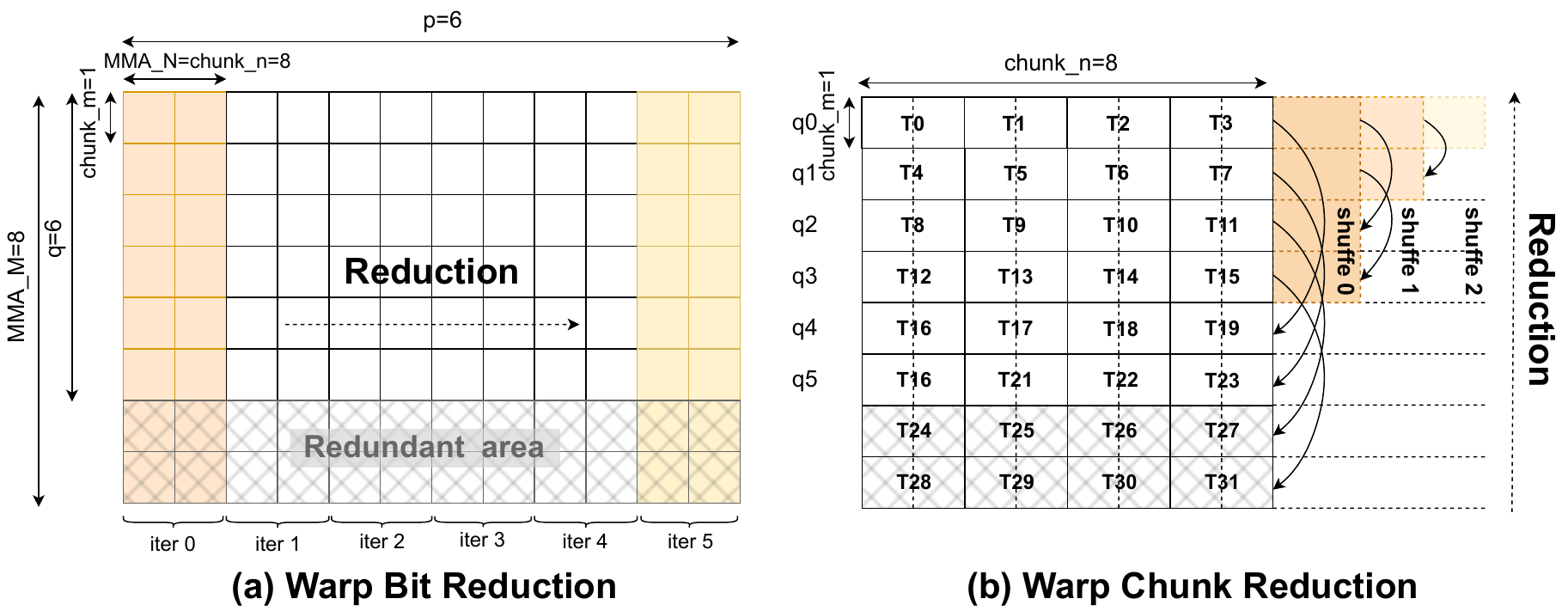}
    \caption{Efficient warp-level reduction, exemplified by GEMV.}
    \label{fig:warp_reduction}
\end{figure}

The reduction stage is the cornerstone of our bit-level GEMM implementation within the FlexQ kernel. It performs multi-level weighted accumulation of intermediate results generated by BMMA operations (Figure \ref{fig:warp_reduction}), which can be systematically divided into \textit{bit-level} and \textit{chunk-level} reduction. The bit-level reduction execution is mandatory and involves summation at binary precision, as described in Equation \ref{con:bitlevelcal}, effectively performing weighted accumulation at the bitwise level. In contrast, the chunk-level reduction is conditionally activated only when the batch size is less than MMA\_M, it aggregates and condenses the bit-level results across larger data segments, operating at a coarser granularity.

To elucidate the reduction process, we consider a representative GEMV scenario with M = chunk\_m = 1. After executing the BMMA stage, a single warp produces an output sub-block $C^t$ of dimensions q*chunk\_m × p*chunk\_n. This sub-block then undergoes the reduction process. Specifically, the bit-level reduction performs weighted summation along the dimensions [WARP\_M\_TILES, WARP\_N\_TILES], with each reduction dimension corresponding to [MMA\_M, MMA\_N]. This reduction results in an intermediate output, $C^w$, which contains the bit-wise weights after aggregation.

Notably, in specific scenarios where the batch size $<$ MMA\_M (i.e., when chunk\_m directly equals the batch size), the bit-level reduction can only perform weighted summation along the WARP\_N\_TILES dimension. In the WARP\_N\_TILES dimension, since chunk\_m $<$ MMA\_M, each group of threads within the warp stores weighted representations of different bits for the same chunk\_m. For example, threads 0–3 might hold the weighted representation of $q^0$ for $X_{0,k}$, threads 4–7 hold the weighted representation of $q^1$ for the same, and so on.  To handle these special cases, an additional chunk-level reduction step is necessary to complement the summation reduction along the WARP\_M\_TILES dimension. This chunk-level reduction involves a total of $\left\lfloor \log_2 (MMA\_M) \right\rfloor-\left\lfloor \log_2 (chunk\_m) \right\rfloor$ shuffle operations. At the implementation level, we utilize NVIDIA GPU-specific \texttt{\_\_shfl\_down\_sync} instructions to enable efficient register-to-register communication. During the $i$-th shuffle, thread $T_j$ retrieves data from thread $T_{j + (\text{warp\_size} / 2^{i+1})}$ (i.e., $T_{j + 2^{4-i}}$) to perform the summation of weighted representations. After completing n shuffles, the first chunk\_m groups of threads have accumulated the weighted sum of each bit for $X_{\text{chunk\_m},k}$. This results in the final output block $C$ with size [chunk\_m, chunk\_n].

Both the two kinds of reductions operate entirely within a warp. From a hardware perspective, this design takes full advantage of the GPU’s SIMT architecture, enabling efficient parallel reduction through close collaboration of the 32 threads in a warp. By leveraging register-shuffle instructions for communication, this approach minimizes costly global or shared memory access, further enhancing efficiency for bit-level GEMM operations.

\subsubsection{Multi-Stage Software Pipeline}
Modern GPU architectures exhibit a pronounced disparity between their high computational throughput and limited memory bandwidth. This mismatch often leads to underutilized compute units, which tend to remain idle while waiting for data from the memory subsystem. Therefore, effectively overlapping memory access with computation is crucial for enhancing GPU kernel performance.

\begin{figure}[ht]
    \centering
    \includegraphics[width=\linewidth]{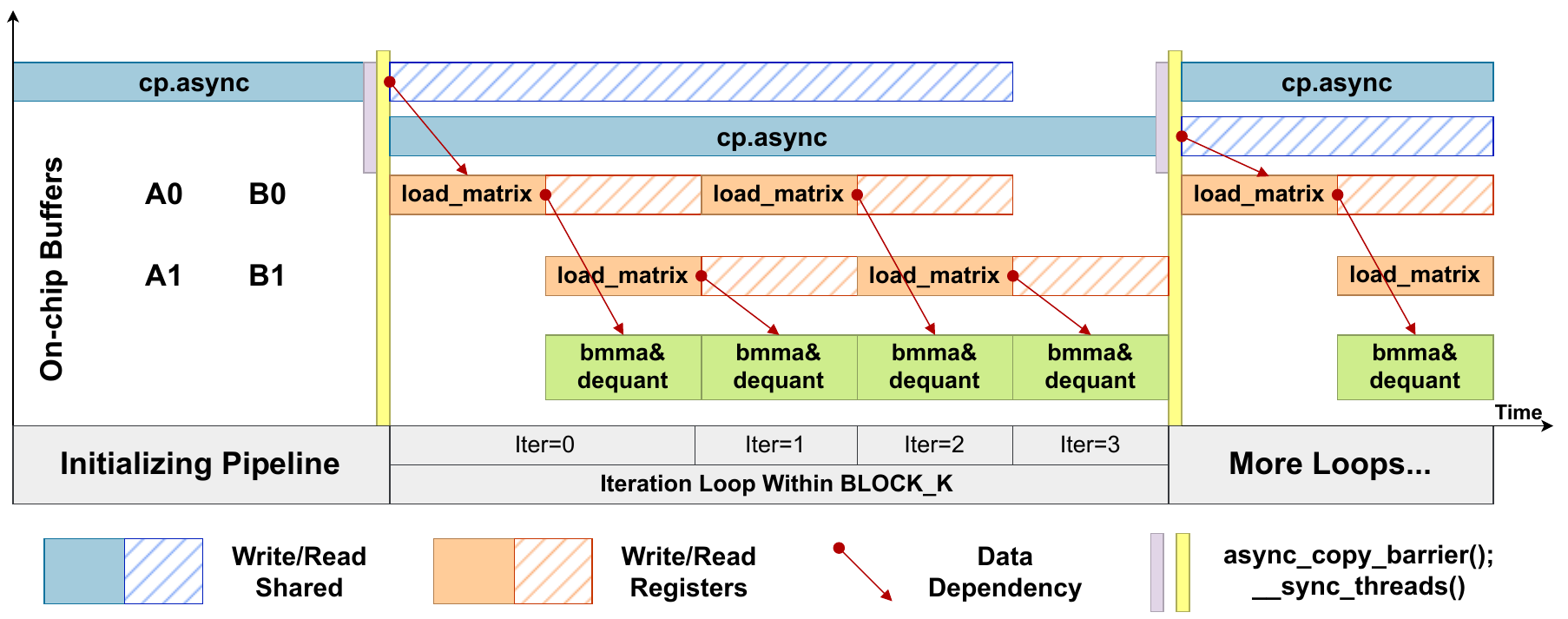}
    \caption{The K-stage software pipeline in FlexQ kernel (illustrated with K = 2).}
    \label{fig:pipeline}
\end{figure}

To this end, we propose a multi-stage software pipelining strategy, as illustrated in Figure \ref{fig:pipeline} (with K = 2). In this approach, matrix multiplication is decomposed into multiple tiles along the inner product dimension. During pipeline initialization, the \texttt{cp.async} instructions introduced with the Ampere architecture are employed to initiate K-1 asynchronous global memory read operations. This prefetching stage retrieves data needed for tiles 0 through K-2, ensuring data readiness exactly when required for computation. Before each tile begins its computation, synchronization barriers guarantee that all necessary data has been written into shared memory. Crucially, this approach allows data loading from global memory for each tile to occur concurrently with the ongoing computations of other tiles, thereby maximizing the overlap between data loading and computation.

Furthermore, we incorporate a double-buffering scheme into the pipeline to facilitate overlapping TC computations (i.e., BMMA operations) and register data transfers. As depicted in Figure \ref{fig:pipeline}, two sets of register buffers, Buffer 0 (A0, B0) and Buffer 1 (A1, B1), are used. At iteration 0, data for Tile 0 is loaded into the first buffer set, and the corresponding BMMA operation begins immediately. Meanwhile, the second buffer set prefetches data for the subsequent iteration from shared memory. At iteration 1, data for this iteration is already loaded into Buffer 1, allowing the BMMA computation to proceed without delay. Simultaneously, data for iteration 2 is prefetched into Buffer 0. This alternating pattern continues iteratively, seamlessly overlapping TC computations with register data transfers.

\section{Evaluation}
\subsection{Experimental Setup}

\begin{table*}[!h]
% \setlength{\tabcolsep}{1mm}
% \small
\begin{center}
\begin{tabular}{c|l|l|cccccccc}
\toprule
\textbf{Dataset}           & \textbf{\#Bit}        & \multicolumn{1}{c|}{\textbf{Method}} & \textbf{1-7B} & \textbf{1-13B} & \textbf{1-30B} & \textbf{1-65B} & \textbf{2-7B} & \textbf{2-13B} & \textbf{2-70B} & \textbf{3-8B} \\ \midrule
\multirow{6}{*}{WikiText2} & FP16                  & -                                    & 5.68          & 5.09           & 4.10           & 3.53           & 5.47          & 4.88           & 3.31           & 6.14           \\ \cmidrule{2-11} 
   & \multirow{5}{*}{W6A6}                         & SmoothQuant                          & 6.03          & 5.42           & 4.55           & 3.88           & 6.20          & 5.18           & 3.69           & 7.07           \\
   &                                               & OmniQuant                            & 5.96          & 5.28           & 4.38           & 3.75           & 5.87          & 5.14           & 3.71           & 7.24           \\
   &                                               & I-LLM                                & 5.84          & 5.23           & 4.32           &  -             & 5.68          & 5.10           &  -             & 6.61           \\
   &                                               & DuQuant                              & 5.74          & 5.13           & 4.15           & 3.60           & 5.53          & 4.92           & 3.35           & 6.27           \\
   &                                               & \cellcolor{gray!30}\textbf{FlexQ}                       & \cellcolor{gray!30}\textbf{5.70} & \cellcolor{gray!30}\textbf{5.12}  & \cellcolor{gray!30}\textbf{4.13}  & \cellcolor{gray!30}\textbf{3.55}  & \cellcolor{gray!30}\textbf{5.52} & \cellcolor{gray!30}\textbf{4.91}  & \cellcolor{gray!30}\textbf{3.34}  & \cellcolor{gray!30}\textbf{6.24}  \\ \midrule
\multirow{6}{*}{C4}        & FP16                  & -                                    & 7.08          & 6.61           & 5.98           & 5.62           & 6.97          & 6.46           & 5.52           & 8.88           \\ \cmidrule{2-11} 
   & \multirow{5}{*}{W6A6}                         & SmoothQuant                          & 7.47          & 6.97           & 6.34           & 5.99           & 7.76          & 6.76           & 5.88           & 9.57           \\
   &                                               & OmniQuant                            & 7.43          & 6.84           & 6.22           & 5.82           & 7.48          & 6.74           & 5.91           & 9.82           \\
   &                                               & I-LLM                                & 7.32          & 6.79           & 6.25           &  -             & 7.27          & 6.74           &  -             & 9.77           \\
   &                                               & DuQuant                              & 7.13          & 6.64           & 6.01           & 5.64           & 7.03          & 6.50           & 5.54           & 9.10           \\
   &                                               & \cellcolor{gray!30}\textbf{FlexQ}                       & \cellcolor{gray!30}\textbf{7.11} & \cellcolor{gray!30}\textbf{6.63}  & \cellcolor{gray!30}\textbf{6.00}  & \cellcolor{gray!30}\textbf{5.63}  & \cellcolor{gray!30}\textbf{7.02} & \cellcolor{gray!30}\textbf{6.49}  & \cellcolor{gray!30}\textbf{5.53}  & \cellcolor{gray!30}\textbf{9.05}  \\
\bottomrule
\end{tabular}
\end{center}
% \caption{Perplexity results of 6-bit weight and activation quantization applied to LLaMA and LLaMA-2 models evaluated on the WikiText2 and C4 datasets.}
\caption{Perplexity results of 6-bit weight and activation quantization applied to LLaMA family models evaluated on the WikiText2 and C4 datasets.}
\label{tab:llama_perplexity}
\end{table*}

% zero-shot llama-1
\begin{table*}[!h]
% \setlength{\tabcolsep}{1mm}
% \small
\begin{center}
\begin{tabular}{c|l|ccccccc}
\toprule
\textbf{Model}     & \textbf{Method}  & \textbf{PIQA}  & \textbf{ARC-E} & \textbf{ARC-C} & \textbf{BoolQ} & \textbf{HellaSwag} & \textbf{WinoGrande} & \textbf{Avg}   \\ \midrule
\multirow{5}{*}{\begin{tabular}[c]{@{}c@{}}LLaMA1-7B\\ W6A6\end{tabular}}  
                   & FP16             & 77.47          & 52.48          & 41.46          & 73.08          & 73.00              & 67.07               & 64.09          \\ \cmidrule{2-9}
                   & SmoothQuant      & 76.75          & 51.64          & 39.88          & 71.75          & 71.67              & 65.03               & 62.81          \\
                   & OmniQuant        & 77.09          & 51.89          & 40.87          & 72.53          & 71.61              & 65.03               & 63.17          \\
                   % & I-LLM            & 76.99          & 52.66          & 40.78          & 72.94          & 71.31              & 65.67               & 63.39          \\
                   & DuQuant          & 77.42          & 52.65          & 40.53          & 71.53          & 72.64              & 67.72               & 63.75          \\ 
                   & \cellcolor{gray!30}\textbf{FlexQ}            & \cellcolor{gray!30}77.09          & \cellcolor{gray!30}53.03          & \cellcolor{gray!30}40.79          & \cellcolor{gray!30}72.78          & \cellcolor{gray!30}72.91              & \cellcolor{gray!30}66.22               & \cellcolor{gray!30}\textbf{63.80} \\ \midrule
\multirow{5}{*}{\begin{tabular}[c]{@{}c@{}}LLaMA1-13B\\ W6A6\end{tabular}} 
                   & FP16             & 79.10          & 59.89          & 44.45          & 68.01          & 76.21              & 70.31               & 66.33          \\ \cmidrule{2-9}
                   & SmoothQuant      & 77.91          & 56.60          & 42.40          & 64.95          & 75.36              & 69.36               & 64.43          \\
                   & OmniQuant        & 78.40          & 57.28          & 42.91          & 67.00          & 75.82              & 68.27               & 64.95          \\
                   % & I-LLM            & 77.49          & 56.94          & 44.03          & 64.92          & 75.24              & 69.14               & 64.63          \\
                   & DuQuant          & 79.16          & 59.39          & 43.69          & 68.10          & 75.81              & 69.06               & 65.87          \\ 
                   & \cellcolor{gray!30}\textbf{FlexQ}            & \cellcolor{gray!30}79.22          & \cellcolor{gray!30}59.18          & \cellcolor{gray!30}44.45          & \cellcolor{gray!30}69.30          & \cellcolor{gray!30}75.96              & \cellcolor{gray!30}70.17               & \cellcolor{gray!30}\textbf{66.38} \\ \midrule
\multirow{5}{*}{\begin{tabular}[c]{@{}c@{}}LLaMA1-30B\\ W6A6\end{tabular}} 
                   & FP16             & 80.08          & 58.92          & 45.47          & 68.44          & 79.21              & 72.53               & 67.44          \\ \cmidrule{2-9}
                   & SmoothQuant      & 77.14          & 57.61          & 42.91          & 65.56          & 78.07              & 69.92               & 65.20          \\
                   & OmniQuant        & 79.81          & 58.79          & 45.22          & 68.38          & 78.95              & 72.21               & 67.23          \\
                   % & I-LLM            & 79.43          & 58.88          & 45.14          & 73.36          & 78.51              & 72.61               & 67.99          \\
                   & DuQuant          & 80.09          & 57.95          & 45.05          & 68.72          & 79.17              & 73.09               & 67.35          \\ 
                   & \cellcolor{gray!30}\textbf{FlexQ}            & \cellcolor{gray!30}80.41          & \cellcolor{gray!30}59.30          & \cellcolor{gray!30}45.22          & \cellcolor{gray!30}69.48          & \cellcolor{gray!30}79.18              & \cellcolor{gray!30}73.17               & \cellcolor{gray!30}\textbf{67.79} \\ \midrule
\multirow{5}{*}{\begin{tabular}[c]{@{}c@{}}LLaMA1-65B\\ W6A6\end{tabular}} 
                   & FP16             & 80.79          & 58.71          & 46.24          & 82.29          & 80.72              & 77.50               & 71.04          \\ \cmidrule{2-9}
                   & SmoothQuant      & 80.25          & 57.92          & 45.50          & 80.22          & 80.18              & 74.76               & 69.80          \\
                   & OmniQuant        & 81.01          & 58.12          & 46.33          & 80.64          & 79.91              & 75.69               & 70.28          \\
                   % & I-LLM            &  -             &  -             &  -             &  -             &  -                 &  -                  &  -             \\
                   & DuQuant          & 80.63          & 58.00          & 46.50          & 82.08          & 80.49              & 76.87               & 70.76          \\ 
                   & \cellcolor{gray!30}\textbf{FlexQ}            & \cellcolor{gray!30}80.79          & \cellcolor{gray!30}58.75          & \cellcolor{gray!30}46.16          & \cellcolor{gray!30}81.84          & \cellcolor{gray!30}80.53              & \cellcolor{gray!30}76.95               & \cellcolor{gray!30}\textbf{70.84} \\ 
\bottomrule
\end{tabular}
\end{center}
\caption{Zero-shot common-sense question answering (QA) results for LLaMA models with 6-bit weight and activation quantization.}
\label{tab:llama1_accuracy}
\end{table*}

\paragraph{Baselines.} 
For the accuracy experiments, we compare our method against SmoothQuant \cite{xiao2023smoothquant}, OmniQuant \cite{shao2024omniquant}, I-LLM \cite{hu2024llm}, and DuQuant \cite{lin2024duquant}. Specifically, for DuQuant, we adopt the DuQuant+LWC quantization model for evaluation. To assess the efficiency of our FlexQ kernels, we compare the W6Ax variant with cuBLAS (W8A8) and ABQ-LLM \cite{zeng2025abq}’s corresponding precision kernels across various GEMM workloads on LLaMA (7B, 30B) \cite{touvron2023llama} and LLaMA-2 (13B, 70B) models \cite{touvron2023llama2}, as well as the OPT-30B model \cite{zhang2022opt}. For end-to-end evaluation, we integrate our approach into FasterTransformer, comparing it against the FP16 baseline, FasterTransformer’s weight-only quantized CUTLASS (W8A16), and the SmoothQuant (W8A8) implementation. It is important to note that ABQ-LLM’s quantization strategy primarily targets extremely low-bit quantization, such as INT2. Since our focus is on INT6 quantization, we exclude ABQ-LLM from the end-to-end evaluation as a baseline.

\paragraph{Workloads.}
% We evaluate FlexQ on several LLMs: LLaMA (7B–65B), LLaMA-2 (7B–70B), and OPT (6.7B–30B). 
We evaluate FlexQ on multiple real-world LLMs, including LLaMA (7B–65B), LLaMA-2 (7B–70B), LLaMA-3 (8B), and OPT (6.7B–30B). Consistent with prior work \cite{xiao2023smoothquant, shao2024omniquant, hu2024llm, lin2024duquant}, we report perplexity scores on language modeling benchmarks, specifically the WikiText2 \cite{merity2017pointer} and C4 \cite{raffel2020exploring} datasets. To evaluate zero-shot learning performance, we adopt a benchmarking methodology similar to previous studies \cite{frantar2022gptq, xiao2023smoothquant, lin2024qserve, liu2025comet}, conducting evaluations on several standard benchmarks utilizing the lm\_eval harness \cite{gao2021framework}, including PIQA \cite{bisk2020piqa}, ARC \cite{clark2018think}, BoolQ \cite{clark2019boolq}, HellaSwag \cite{zellers2019hellaswag}, and Winogrande \cite{sakaguchi2021winogrande}.

\paragraph{Implementation Details.} 
For accuracy experiments, we evaluate FlexQ implemented in PyTorch using the HuggingFace Transformers library \cite{wolf2019huggingface}. We employ a group size of 128, applying 6-bit fine-grained group-wise symmetric quantization uniformly to all weight matrices. For activation quantization, we use 8-bit precision for the critical quantization-sensitive down\_proj layer, while applying 6-bit quantization to other linear layers. Since OPT models are based on a non-GLU architecture, we quantize both weights and activations to 6 bits uniformly. To optimize execution efficiency, we utilize the Auto Kernel Search from ABQ-LLM to determine the best kernel block size and configuration parameters tailored to different computational shapes.

% zero-shot llama-2
\begin{table*}[h]
% \setlength{\tabcolsep}{1mm}
% \small
\begin{center}
\begin{tabular}{c|l|ccccccc}
\toprule
\textbf{Model}     & \textbf{Method}  & \textbf{PIQA} & \textbf{ARC-E} & \textbf{ARC-C} & \textbf{BoolQ} & \textbf{HellaSwag} & \textbf{WinoGrande} & \textbf{Avg}    \\ \midrule
\multirow{5}{*}{\begin{tabular}[c]{@{}c@{}}LLaMA2-7B\\ W6A6\end{tabular}}  
                   & FP16             & 76.88         & 53.54          & 40.53          & 71.13          & 72.96              & 67.25               & 63.72            \\ \cmidrule{2-9} 
                   & SmoothQuant      & 75.57         & 53.62          & 39.93          & 69.54          & 71.76              & 66.14               & 62.76            \\
                   & OmniQuant        & 76.55         & 53.83          & 40.96          & 68.75          & 55.89              & 65.59               & 60.26            \\
                   % & I-LLM            &               &                &                &                &                    &                     &                  \\
                   & DuQuant          & 76.88         & 52.31          & 40.44          & 69.72          & 72.60              & 66.93               & 63.15            \\
                   & \cellcolor{gray!30}\textbf{FlexQ}            & \cellcolor{gray!30}77.10         & \cellcolor{gray!30}53.33          & \cellcolor{gray!30}41.30          & \cellcolor{gray!30}69.82          & \cellcolor{gray!30}72.78              & \cellcolor{gray!30}67.64               & \cellcolor{gray!30}\textbf{63.66}   \\ \midrule
\multirow{5}{*}{\begin{tabular}[c]{@{}c@{}}LLaMA2-13B\\ W6A6\end{tabular}} 
                   & FP16             & 79.05         & 57.91          & 44.20          & 69.02          & 76.60              & 69.69               & 66.08            \\ \cmidrule{2-9} 
                   & SmoothQuant      & 78.29         & 57.41          & 43.86          & 69.50          & 75.02              & 66.93               & 65.17            \\
                   & OmniQuant        & 78.24         & 57.58          & 43.86          & 71.10          & 75.52              & 68.35               & 65.78            \\
                   % & I-LLM            &               &                &                &                &                    &                     &                  \\
                   & DuQuant          & 78.94         & 57.95          & 44.11          & 68.81          & 76.17              & 68.98               & 65.83            \\ 
                   & \cellcolor{gray!30}\textbf{FlexQ}            & \cellcolor{gray!30}79.16         & \cellcolor{gray!30}58.00          & \cellcolor{gray!30}43.86          & \cellcolor{gray!30}68.29          & \cellcolor{gray!30}76.37              & \cellcolor{gray!30}69.69               & \cellcolor{gray!30}\textbf{65.90}   \\ \midrule
\multirow{5}{*}{\begin{tabular}[c]{@{}c@{}}LLaMA2-70B\\ W6A6\end{tabular}} 
                   & FP16             & 81.01         & 59.68          & 47.95          & 75.87          & 80.87              & 76.95               & 70.39            \\ \cmidrule{2-9} 
                   & SmoothQuant      & 79.87         & 57.32          & 45.65          & 77.13          & 79.01              & 74.03               & 68.84            \\
                   & OmniQuant        & 80.20         & 60.27          & 46.84          & -              & 80.55              & 76.01               & 68.77            \\
                   % & I-LLM            &               &                &                &                &                    &                     &                  \\
                   & DuQuant          & 81.18         & 59.26          & 47.78          & 77.86          & 80.68              & 76.95               & \textbf{70.62}   \\ 
                   & \cellcolor{gray!30}\textbf{FlexQ}            & \cellcolor{gray!30}80.85         & \cellcolor{gray!30}59.76          & \cellcolor{gray!30}48.46          & \cellcolor{gray!30}77.34          & \cellcolor{gray!30}80.72              & \cellcolor{gray!30}76.40               & \cellcolor{gray!30}70.59            \\ \midrule
\multirow{5}{*}{\begin{tabular}[c]{@{}c@{}}LLaMA3-8B\\ W6A6\end{tabular}} 
                   & FP16             & 80.85         & 77.78          & 53.41          & 81.28          & 79.16              & 72.84               & 74.22            \\ \cmidrule{2-9} 
                   & SmoothQuant      & 78.94         & 75.88          & 49.49          & 77.58          & 77.39              & 70.80               & 71.68            \\
                   & OmniQuant        & 78.90         & 73.95          & 47.35          & 74.95          & 76.77              & 70.56               & 70.41            \\
                   % & I-LLM            &               &                &                &                &                    &                     &                  \\
                   & DuQuant          & 79.71         & 77.57          & 53.07          & 80.00          & 78.70              & 73.09               & 73.69            \\ 
                   & \cellcolor{gray!30}\textbf{FlexQ}            & \cellcolor{gray!30}80.52         & \cellcolor{gray!30}78.16          & \cellcolor{gray!30}53.16          & \cellcolor{gray!30}81.35          & \cellcolor{gray!30}78.79              & \cellcolor{gray!30}73.64               & \cellcolor{gray!30}\textbf{74.27}   \\
\bottomrule
\end{tabular}
\end{center}
% \caption{Zero-shot common-sense QA performance of LLaMA2 models using 6-bit weight and activation quantization.}
\caption{Zero-shot common-sense QA performance of LLaMA-2 and LLaMA-3 models using 6-bit weight and activation quantization.}
\label{tab:llama2&3_accuracy}
\end{table*}

\subsection{Accuracy Evaluation}

\paragraph{LLaMA Family Perplexity.} 
Table \ref{tab:llama_perplexity} reports the perplexity results of FlexQ and leading baselines on the LLaMA family models. FlexQ delivers competitive results on both WikiText2 and C4, consistently outperforming SmoothQuant, OmniQuant, and I-LLM across all model sizes. Relative to the current state-of-the-art DuQuant, FlexQ is either on par or superior, trimming WikiText2 perplexity by 0.01–0.1. Remarkably, FlexQ achieves nearly identical performance to FP16 baselines while being calibration-free, offering an almost lossless and highly efficient deployment solution for LLMs.

\paragraph{Zero-shot Accuracy.} 
To further validate our method, we compare zero-shot accuracy of FlexQ against selected baselines across six common-sense tasks, as shown in Tables \ref{tab:llama1_accuracy} and \ref{tab:llama2&3_accuracy}. FlexQ outperforms the baselines on average across all LLaMA models of different scales. Remarkably, for certain tasks, FlexQ even surpasses the accuracy of the FP16 baseline, highlighting its ability to preserve the generalization capabilities of LLM post-quantization.

\begin{table}[!h]
    \setlength{\tabcolsep}{1mm}
    \small
    \center
    \begin{tabular}{cccccccc}
    \toprule
    \multirow{2}[3]{*}{\#Bits} & \multirow{2}[3]{*}{Method} & \multicolumn{2}{c}{OPT-6.7B} & \multicolumn{2}{c}{OPT-13B} & \multicolumn{2}{c}{OPT-30B} \\
    \cmidrule(lr){3-4}\cmidrule(lr){5-6}\cmidrule(lr){7-8}&       & Wiki & C4    & Wiki & C4    & Wiki & C4 \\ \midrule
        FP16    & -             & 10.86  & 11.74     & 10.13  & 11.20     & 9.56   & 10.69  \\ \midrule
    \multirow{4}[2]{*}{W6A6} 
                & SmoothQuant   & 11.34          & 12.14          & 10.56          & 11.40          & 9.67          & 10.81           \\
                & OmniQuant     & 10.96          & \textbf{11.81} & 10.21          & \textbf{11.27} & 9.62          & 10.76           \\
                & I-LLM         & 10.94          & 11.82          & \textbf{10.17} & 11.90          & 9.72          & 10.83           \\
                & \cellcolor{gray!30}\textbf{FlexQ}         & \cellcolor{gray!30}\textbf{10.91} & \cellcolor{gray!30}11.87          & \cellcolor{gray!30}10.37          & \cellcolor{gray!30}\textbf{11.27} & \cellcolor{gray!30}\textbf{9.60} & \cellcolor{gray!30}\textbf{10.75}  \\
    \bottomrule
    \end{tabular}%
    \caption{Perplexity results for OPT models with 6-bit weight and activation quantization on the WikiText2 (denoted as Wiki) and C4 datasets.}
    \label{tab:opt_perplexity}
\end{table}

\paragraph{OPT Family Perplexity.} 
To evaluate FlexQ’s applicability across diverse LLM architectures, we further assess it on OPT models, as detailed in Table \ref{tab:opt_perplexity}. Since DuQuant was not tested on OPT models, it is excluded as a baseline here. FlexQ consistently outperforms previous methods in most cases, maintaining excellent accuracy even on non-GLU architectures. This evidences FlexQ's robust applicability to diverse architectures.

\begin{figure*}[!ht]
    \centering
    \includegraphics[width=\textwidth]{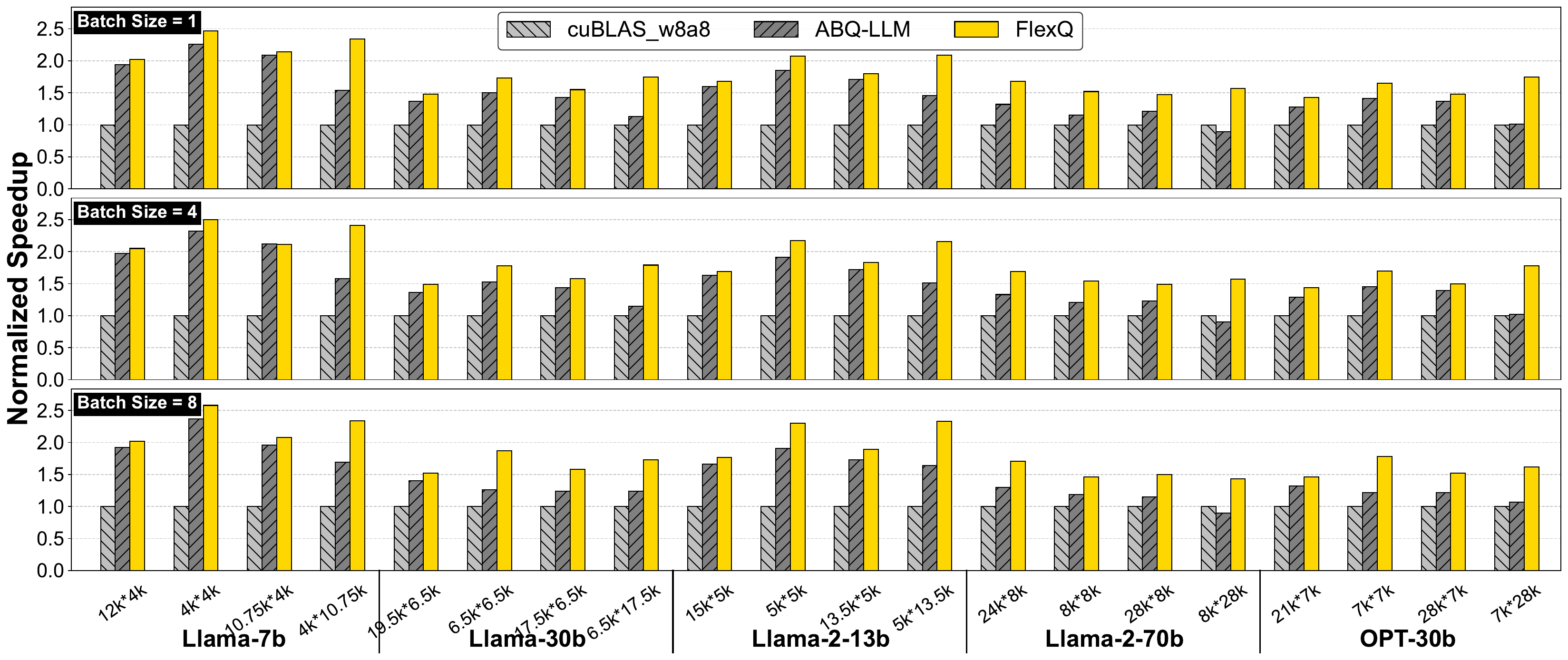}
    \caption{Speedups of linear layers relative to baseline methods during the token generation phase on an NVIDIA RTX 3090 Ti GPU.}
    \label{fig:kernel_benchmark}
\end{figure*}

\begin{figure*}[!h]
    \centering
    \includegraphics[width=\linewidth]{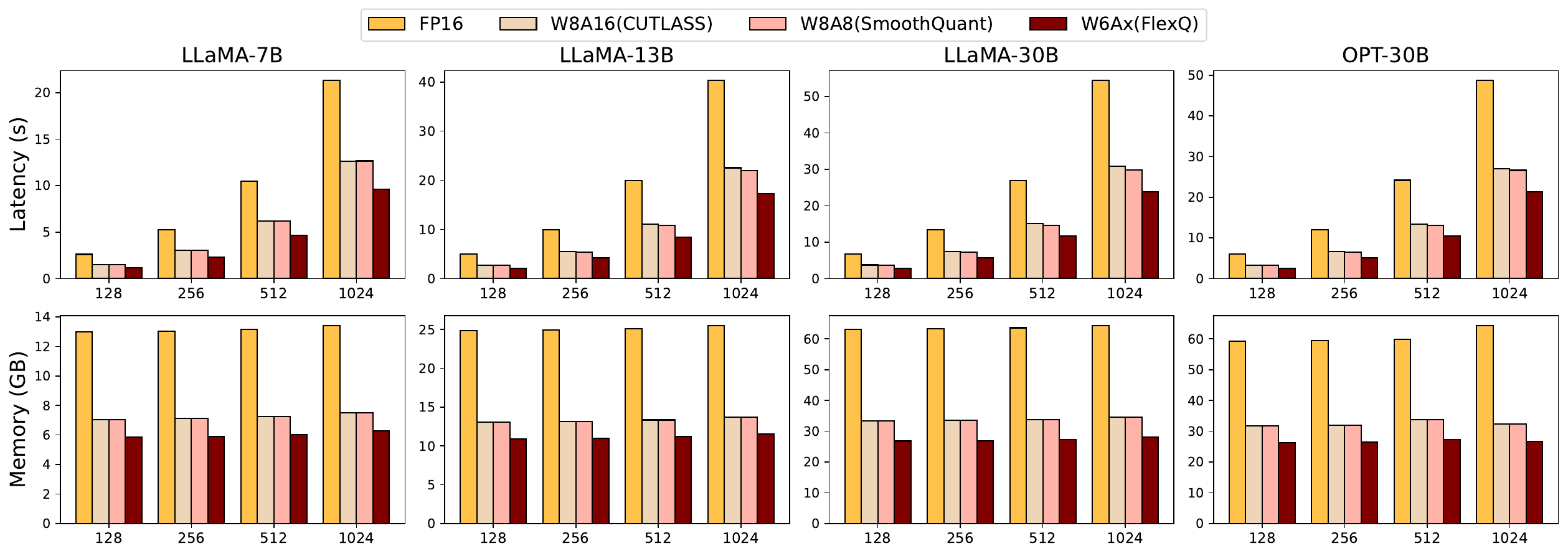}
    \caption{Inference latency (top) and memory utilization (bottom) measurements for the FasterTransformer implementation, evaluated on NVIDIA A6000-48GB GPU across different sequence lengths, with a comparative analysis at a fixed input length of 15 tokens.}
    \label{fig:same_seq_len_e2e}
\end{figure*}

\begin{figure}[!h]
    \centering
    \includegraphics[width=\linewidth]{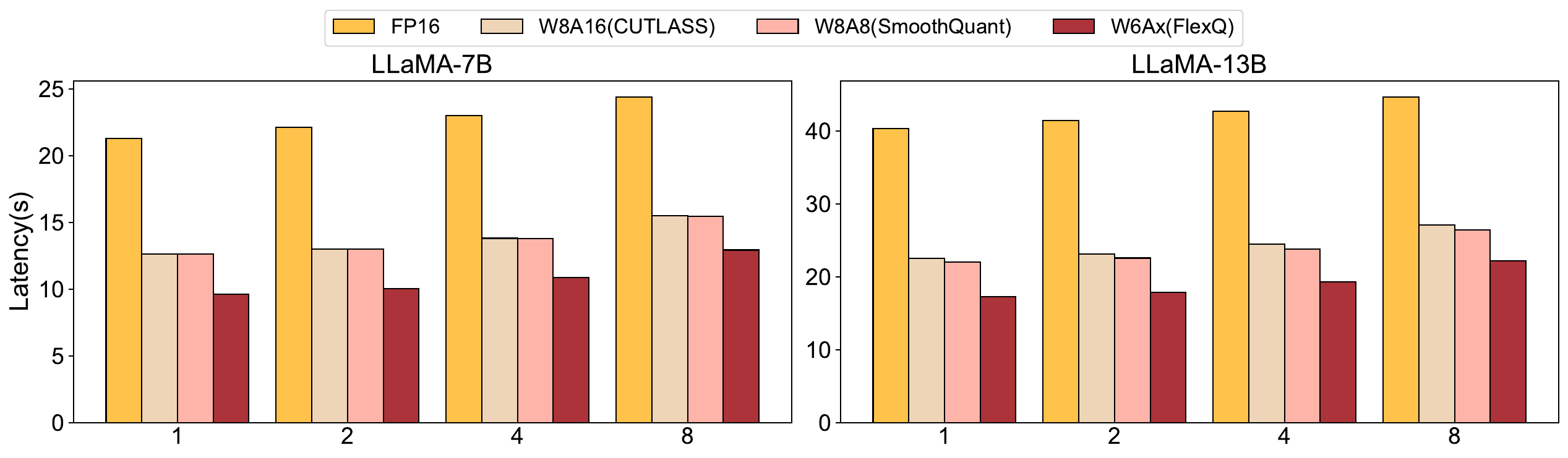}
    \caption{Comparison of same-batch latency between FlexQ and the baseline for LLaMA-7B and LLaMA-13B on an NVIDIA A6000-48GB GPU, using a sequence length of 1024.}
    \label{fig:same_batch_size_e2e}
\end{figure}

\subsection{Efficiency Evaluation}
We evaluate the efficiency of FlexQ through comprehensive experiments measuring both the kernel-level performance of the W6A8 kernel and the overall end-to-end inference throughput of LLMs. All evaluations were conducted on NVIDIA RTX 3090 Ti 24GB and A6000 48GB GPUs, utilizing CUDA 12.1.

\paragraph{Kernel Performance.} \label{sec:Kernel_Performance}
We benchmark the FlexQ W6Ax kernel across various GEMM workloads representative of different-scale LLaMA and OPT models. Experiments employ batch sizes of 1, 4, and 8 to assess the kernel’s adaptability to diverse inference scenarios, with particular focus on the critical GEMV operations. For comparison, we include the cuBLAS W8A8 quantized kernel and the ABQ-LLM W6A6/W6A8 kernels. Notably, due to the design of the quantization scheme, both ABQ-LLM and FlexQ employ the W6A8 kernel specifically for the ``down\_proj" GEMM, which is highly sensitive to quantization accuracy.

Figure \ref{fig:kernel_benchmark} presents the speedup of the FlexQ W6Ax kernel relative to baseline kernels. Results demonstrate that FlexQ consistently outperforms the baselines across a wide range of matrix sizes relevant to LLM inference. Interestingly, the ABQ-LLM W6A8 kernel exhibits subpar performance on certain workloads, most notably on the matrix dimensions (1, 28672) × (28672, 8192), where its computational throughput falls below that of cuBLAS. In contrast, FlexQ maintains superior performance across all tested workloads. Specifically, with batch sizes of 1, 4, and 8, FlexQ achieves average speedups of 1.78×, 1.81×, and 1.82× over cuBLAS, and 1.24×, 1.24×, and 1.27× over ABQ-LLM, respectively.

\paragraph{End-to-End Evaluation.}  \label{sec:E2E_Evaluation}
To assess FlexQ’s performance in end-to-end inference scenarios, we integrate the FlexQ W6Ax kernel into FasterTransformer and compare it against multiple baselines, including the FP16 implementation of FasterTransformer, the W8A16 implementation based on Cutlass, and the W8A8 version of SmoothQuant. As shown in Figure \ref{fig:same_seq_len_e2e}, results on the LLaMA-13B model demonstrate that FlexQ achieves up to 2.38× inference acceleration and 2.28× memory compression relative to FP16. Across the LLaMA and OPT model families, FlexQ consistently surpasses mainstream inference methods, delivering 1.25–1.33× speedup and 1.19–1.24× reduction in memory footprint compared to SmoothQuant. This improvement primarily stems from FlexQ’s full utilization of BTCs, which leverages the potential of 6-bit quantization to achieve higher computational efficiency.

Additionally, we present inference acceleration results under identical batch sizes in Figure \ref{fig:same_batch_size_e2e}. FlexQ outperforms all other quantization approaches across all batch sizes tested. Notably, at a batch size of 8, FlexQ achieves up to 1.89× and 2.01× inference speedup over FP16 on LLaMA-7B and LLaMA-13B models, respectively. Under varying batch sizes, FlexQ achieves 1.2–1.32× and 1.19–1.28× acceleration relative to SmoothQuant for LLaMA-7B and LLaMA-13B, respectively. These end-to-end results demonstrates that the proposed 6-bit FlexQ quantization method offers substantial performance advantages over existing 8-bit quantization techniques, presenting a more practical solution for the efficient deployment of LLMs.

\begin{table}[h]
    \small
    \centering
    \begin{tabular}{ccc}
        \toprule
        Method                          & Latency(us)   $\downarrow $ & TOPS $\uparrow $ \\ \midrule
        cuBLAS                          & 37.02                       & 0.91             \\ \midrule
        Vanilla  Kernel                   & 29.44                       & 1.14             \\
        + Warp-Level Reduction          & 26.54                       & 1.26             \\
        + Multi-Stage Software Pipeline & 16.12                       & 2.08             \\
        + Cache Eviction                & \textbf{15.03}              & \textbf{2.23}    \\
        - Fused Dequantization          & 14.92                       & 2.25             \\ \bottomrule
    \end{tabular}
    \caption{Ablation study examining the impact of various optimization techniques on latency and throughput of the FlexQ kernel, conducted on an RTX 3090 Ti.}
    \label{tab:kernel_ablation}
\end{table}

\subsection{Ablation Study}
To evaluate the contributions of various optimization techniques and fused dequantization operations on kernel performance improvement, we conduct an ablation study on the FlexQ W6A6 kernel. Table \ref{tab:kernel_ablation} presents results for the key GEMV operations with a computational dimension of (1, 4096) × (4096, 4096), where the Auto Kernel Search method is first applied to the vanilla kernel implementation\footnote{The vanilla kernel refers to the initial equivalent BTC-based INT6 implementation that does not incorporate the additional optimizations outlined in Table \ref{tab:kernel_ablation}.}. Without the warp reduction optimization, warp-level intermediate results are stored in shared memory and subsequently read to complete the reduction. Leveraging lower bitwidth weights and BTCs with higher peak throughput, our vanilla kernel implementation already significantly outperforms the cuBLAS W8A8 kernel. With the adoption of additional optimization techniques, such as cache eviction, warp reduction, and pipelining, our final W6A6 kernel achieves a 2.46× reduction in inference latency and a corresponding 2.46× throughput boost, outperforming the cuBLAS baseline by a substantial margin. Notably, for this GEMV scenario, the overhead introduced by fused dequantization is less than 1\%, resulting in negligible performance loss.

\section{Conclusion}
This paper introduces FlexQ, an algorithm-system co-designed framework for LLM inference tailored to INT6 quantization. At the algorithmic level, we integrate fine-grained group quantization with quantization-sensitive layer mixed-precision strategies, achieving negligible accuracy loss under 6-bit quantization. At the system level, we develop the FlexQ W6Ax kernel, which exploits BTCs and introduces memory-efficient bit-level data layouts alongside a suite of novel optimization techniques. FlexQ can be seamlessly integrated into existing inference systems, enabling end-to-end 6-bit support for quantized LLM inference and offering a superior trade-off between model quality and inference efficiency. Extensive experiments demonstrate that FlexQ achieves accuracy comparable to FP16 baselines. In addition, the FlexQ kernel delivers up to 1.82× and 1.27× throughput improvements over cuBLAS and ABQ-LLM, respectively, while reducing end-to-end inference latency by 1.33× compared to the state-of-the-art baseline, SmoothQuant. These results validate FlexQ as a practical and highly efficient solution for low-bit quantized LLM inference.

\bibliography{aaai2026}

\end{document}